\documentclass[sigconf]{aamas}

\usepackage{balance} 
\usepackage{microtype}
\usepackage{graphicx}
\usepackage{subfig}
\usepackage{booktabs} 

\usepackage{amsmath}
\usepackage{mathtools}

\usepackage[utf8]{inputenc} 
\usepackage[T1]{fontenc}    
\usepackage{hyperref}       
\usepackage{url}            
\usepackage{booktabs}       
\usepackage{amsfonts}       
\usepackage{nicefrac}       
\usepackage{microtype}      
\usepackage{xcolor}         
\usepackage{pifont}

\usepackage{diagbox}
\usepackage{multirow}
\usepackage{booktabs}
\usepackage[font={small},labelfont={bf}]{caption}
\usepackage{bbm}
\usepackage{bm}
\usepackage{float}
\usepackage{cite}
\usepackage{wrapfig}
\usepackage{dirtree}

\usepackage[scaled=0.85]{DejaVuSansMono} 

\usepackage{algorithmic, algorithm}

\usepackage{paralist}

\newcommand{\task}{\textit{talking vehicles}}
\newcommand{\simulator}{\text{{TalkingVehiclesGym}}}
\newcommand{\method}{\textsc{CoopReflect}}

\definecolor{asparagus}{rgb}{0.53, 0.66, 0.42}
\definecolor{bittersweet}{rgb}{1.0, 0.44, 0.37}
\definecolor{ao(english)}{rgb}{0.0, 0.5, 0.0}
\definecolor{mydarkblue}{rgb}{0,0.08,0.45}
\definecolor{blue(ncs)}{rgb}{0.0, 0.53, 0.74}
\definecolor{celestialblue}{rgb}{0.29, 0.59, 0.82}
\definecolor{earthyellow}{rgb}{0.88, 0.66, 0.37}
\definecolor{lightgray}{rgb}{0.83, 0.83, 0.83}
\definecolor{brickred}{rgb}{0.8, 0.25, 0.33}
\definecolor{myblue}{rgb}{0.345, 0.545, 0.902}

\usepackage{natbib}
\usepackage{hyperref}
\usepackage[capitalize,noabbrev]{cleveref}

\ifdefined\nohyperref\else\ifdefined\hypersetup
  \hypersetup{ %
    pdftitle={},
    pdfsubject={},
    pdfkeywords={},
    pdfborder=0 0 0,
    pdfpagemode=UseNone,
    colorlinks=true,
    linkcolor=mydarkblue,
    citecolor=ao(english),
    filecolor=mydarkblue,
    urlcolor=magenta,
    }
  \fi
\fi

\newcommand{\secref}[1]{\underline{\hyperref[#1]{Section~\ref*{#1}}}}
\newcommand{\figref}[1]{\underline{\hyperref[#1]{Figure~\ref*{#1}}}}
\newcommand{\tabref}[1]{\underline{\hyperref[#1]{Table~\ref*{#1}}}}
\newcommand{\chapref}[1]{\underline{\hyperref[#1]{Chapter~\ref*{#1}}}}
\newcommand{\eqnref}[1]{\underline{\hyperref[#1]{Equation~\ref*{#1}}}}
\newcommand{\algoref}[1]{\underline{\hyperref[#1]{Algorithm~\ref*{#1}}}}
\newcommand{\appref}[1]{\underline{\hyperref[#1]{Appendix~\ref*{#1}}}}
\usepackage[font={small},labelfont={bf}]{caption}

\doi{MOAV6406}

\makeatletter
\gdef\@copyrightpermission{
  \begin{minipage}{0.2\columnwidth}
   \href{https://creativecommons.org/licenses/by/4.0/}{\includegraphics[width=0.90\textwidth]{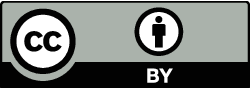}}
  \end{minipage}\hfill
  \begin{minipage}{0.8\columnwidth}
   \href{https://creativecommons.org/licenses/by/4.0/}{This work is licensed under a Creative Commons Attribution International 4.0 License.}
  \end{minipage}
  \vspace{5pt}
}
\makeatother

\setcopyright{ifaamas}
\acmConference[AAMAS '26]{Proc.\@ of the 25th International Conference
on Autonomous Agents and Multiagent Systems (AAMAS 2026)}{May 25 -- 29, 2026}
{Paphos, Cyprus}{C.~Amato, L.~Dennis, V.~Mascardi, J.~Thangarajah (eds.)}
\copyrightyear{2026}
\acmYear{2026}
\acmDOI{}
\acmPrice{}
\acmISBN{}

\acmSubmissionID{374}

\title{\method{}: Towards Natural Language Communication for Cooperative Autonomous Driving via Multi-Agent Learning}

\author{Jiaxun Cui}
\affiliation{
    \institution{The University of Texas at Austin}  
    \city{}
    \country{}
}
\email{cuijiaxun@utexas.edu}
\author{Chen Tang}
\affiliation{
    \institution{University of California, Los Angeles}
    \city{}
    \country{}
}
\email{ctangac@ucla.edu}
\author{Jarrett Holtz}
\affiliation{
    \institution{Robert Bosch LLC}
    \city{}
    \country{}
}
\email{jarrett.holtz@us.bosch.com}
\author{Janice Nguyen}
\affiliation{
    \institution{University of California, Riverside}
    \city{}
    \country{}
}
\email{jnguy172@ucr.edu}
\author{Alessandro G. Allievi}\affiliation{
    \institution{Robert Bosch LLC}
    \city{}
    \country{}
}
\email{alessandro.allievi@us.bosch.com}
\author{Hang Qiu}
\affiliation{
    \institution{University of California, Riverside}
    \city{}
    \country{}
}
\email{hangq@ucr.edu}
\author{Peter Stone}
\affiliation{
    \institution{The University of Texas at Austin \& Sony AI}  
    \city{}
    \country{}
}
\email{pstone@cs.utexas.edu}
\begin{abstract}
Past work has demonstrated that autonomous vehicles can drive more safely if they communicate with each other. 
However, this communication is usually not human-understandable. 
Using natural language as a vehicle-to-vehicle (V2V) communication protocol offers the potential for autonomous vehicles to drive cooperatively not only with each other but also with human drivers. 
To explore the potential use of natural language for V2V communication, we develop LLM-based driving agents and study their interactions in a new simulation environment, \simulator{}, which features traffic scenarios where communication can potentially help avoid imminent collisions and/or support efficient traffic flow.
While LLM agents relying solely on chain-of-thought reasoning struggle to coordinate effectively, we introduce \method{}, a multi-agent learning framework that equips agents with knowledge for both natural language message generation and high-level decision-making through trial and error and multi-agent debriefing. Experiments show that \method{} produces more meaningful and human-understandable messages than existing baselines, enabling stronger cooperation. Finally, we distill scenario-specific knowledge into a unified language model policy, achieving cross-scenario generalization and substantially reducing decision-making latency.
Our code and demo videos are available at 
\url{https://talking-vehicles.github.io/}
\end{abstract}

\keywords{V2V Communication, Learning LLM Agents, Autonomous Driving}

\newcommand{\BibTeX}{\rm B\kern-.05em{\sc i\kern-.025em b}\kern-.08em\TeX}

\begin{document}

\pagestyle{fancy}
\fancyhead{}

\maketitle 

\section{Introduction}
Driving is inherently a multi-agent problem~\citep{dinneweth2022multi, seff2023motionlm}, in which each driver makes independent decisions based on their own perceptions while interacting with others on the road. As we transition towards \mbox{(semi-)}autonomous vehicles, centralized control~\citep{marl2022aim} of all cars may appear efficient, but it is impractical and unlikely to gain widespread adoption. On the other hand, cooperative driving through communication channels is feasible and can offer significant benefits even when implemented in a limited capacity.
Past research has demonstrated the advantages of cooperation among autonomous cars for perception~\citep{wang2020v2vnet,  xu2022opv2v}, prediction~\citep{wang2025cmp}, and planning~\citep{cui2022coopernaut}. However, these benefits are limited to vehicles that use the same learned environmental representation and communication language, limiting broader participation from those with different representations or language and leaving human drivers reliant solely on their local perceptions without being privy to the collaboration efforts.

As vision-language models become increasingly prevalent for language-conditioned reasoning and high-level planning in complex traffic environments, the use of natural language as a complementary, general-purpose communication interface offers significant potential for both vehicle–human and vehicle–vehicle coordination. 
Prior work has made progress toward this vision by training driving agents to generate and explain driving decisions in natural language \citep{wayve2023lingo1, ma2023dolphins} or to coordinate with human drivers within a single vehicle \citep{deruyttere2022talk2car}, often leveraging large-scale datasets \citep{kim2018bddx, kim2019HAD, qian2023nuscenesqa, sima2023drivelm}.
However, \textbf{inter-vehicle} communication using natural language is relatively underexplored, 
particularly with respect to its feasibility and design considerations in cooperative driving scenarios, 
despite a few contemporaneous efforts~\citep{gao2025langcoop,hu2024agentscodriver}. These efforts demonstrate a growing interest in the problem but remain preliminary in scope and evaluation.
Complementing and extending this contemporaneous work, we introduce \textbf{\simulator{}}, a multi-agent simulation framework that models vehicle-to-vehicle communication and enables closed-loop evaluations of natural language interactions across a suite of accident-prone traffic scenarios.

Recent advances in Large Language Models (LLMs) present new opportunities for agents to learn to speak and understand natural language messages in cooperative driving scenarios. In this work, we study how LLM agents can interact using natural language and optimize communication strategies through trial-and-error multi-agent interactions. Our initial experiments show that LLM agents relying only on chain-of-thought reasoning struggle to perform well,  and single-agent reflection methods provide only modest improvements in coordination. Hence, we introduce \textbf{\method{}}, a multi-agent learning method enabling LLM agents to engage in centralized reflection to refine their cooperation strategies. The resulting reflections are later incorporated as memories into decentralized agent execution. Our experimental results in simulation indicate that, when LLM agents initially fail to collaborate effectively, our proposed learning method helps them both learn what to communicate and how to respond to messages through interactions. Finally, we distill the learned behaviors of large models in each scenario into a compact language model, achieving scenario and role generalization, as well as near-real-time inference with decision latency under 500~ms, compared to the 10~s required by large models in wall-clock time.

In summary, our contributions are threefold:
\begin{enumerate}
    \item We introduce \simulator{}, a multi-agent simulation environment that enables closed-loop evaluation of natural language communication in cooperative driving scenarios;
    \item We propose \method{}, a multi-agent learning framework that enables LLM agents to refine cooperation strategies through centralized reflection over multi-agent interactions;
    \item We demonstrate that the cooperative strategies learned by \method{} can be distilled into a single compact language model, achieving efficient inference and generalization.
\end{enumerate}

While this exploratory work does not attempt to address the challenges required to make it fully human-usable --- e.g., by enforcing short, real-time messaging --- this paper takes a crucial step in that direction by restricting all messages to be in natural language and providing a testbed that allows closed-loop improvements. %
\section{Problem Definition}
\label{section:problem}

In this paper, we focus on the subset of agents that actively participate in the cooperation. We assume that these cooperative vehicles implicitly aim to help each other, treating all other (referred to as "background") vehicles as uncontrollable elements of the environment.
Therefore, we frame the problem of \textit{\textbf{Talking Vehicles}} as a partially observable stochastic game (POSG), focusing on optimizing the social welfare of a 
\emph{focal population} ($\mathcal{F}$) \citep{agapiou2022melting} — defined as the joint reward of all participating agents — as the primary objective. 
The reward functions associated with each agent's individual tasks may or may not fully align, necessitating coordination among agents to achieve high joint rewards.
Each agent's observation space is limited to a partial view of the full state, and agents make decisions in a decentralized manner based on their own partial observations and messages received from other agents. In this problem, each agent's action space comprises two main components: \textbf{(1) message generation} and \textbf{(2) vehicle control}. In this work, the message generation space is \emph{open-vocabulary} over natural language (English), instead of being restricted to predefined templates.

We define a POSG with a deterministic observation function and undiscounted rewards as the tuple $$\langle \mathcal{I}, \mathcal{S}, \{\mathcal{O}_i\}, \{\mathcal{A}_i\}, \mathcal{P}, \{\mathcal{R}_i\}\rangle,$$
where $\mathcal{I}=\{1,2,..., N\}$ refers to the identities of all agents in a scenario; 
$\mathcal{S}$ is the state space comprehensively describing the environment; $\mathcal{O}_i$ is the observation space describing agent $i$'s view of the state; $\mathcal{A}_i$ is the action space of agent $i$; $\mathcal{P}$ is the state transition function $\mathcal{S}\times\mathcal{A}_1\times\mathcal{A}_2\times...\times\mathcal{A}_N\rightarrow\mathcal{S}$; $\mathcal{R}_i$ is the reward function of agent $i$.
The focal group of agents is denoted by $\mathcal{F}\subseteq\mathcal{I}$, representing a subset of all agents $\mathcal{I}$. Here, an agent refers to an entity in the POSG (\textit{e.g.}, a vehicle), controlled by a policy $\pi_i$ that specifies how agent $i$ selects actions based on its available information. The goal of each agent $i\in \mathcal{F}$ is to learn a policy $\pi_i$ to maximize the expected cumulative task returns of all agents in $\mathcal{F}$, given background agent policies outside the focal group:
 $
\textcolor{myblue}{\max_{\{\pi_i\}_{i\in{\mathcal{F}}}}~\mathbb{E}\Big[\sum_{{i\in{\mathcal{F}}}} \sum_{t=0}^{t=T} {{{R_i}(s_t, \mathbf{a}_t)}} \Big| {\{\pi_{{j}}\}}_{{j\notin\mathcal{F}, j\in\mathcal{I}}}\Big]}
$,
where $s_t$ is the state at time $t$, and $\mathbf{a_t}=(a_1^t, a_2^t, ..., a_N^t)$ is the joint action of all agents at time $t$.

\begin{figure}
    \centering
    \includegraphics[width=\linewidth]{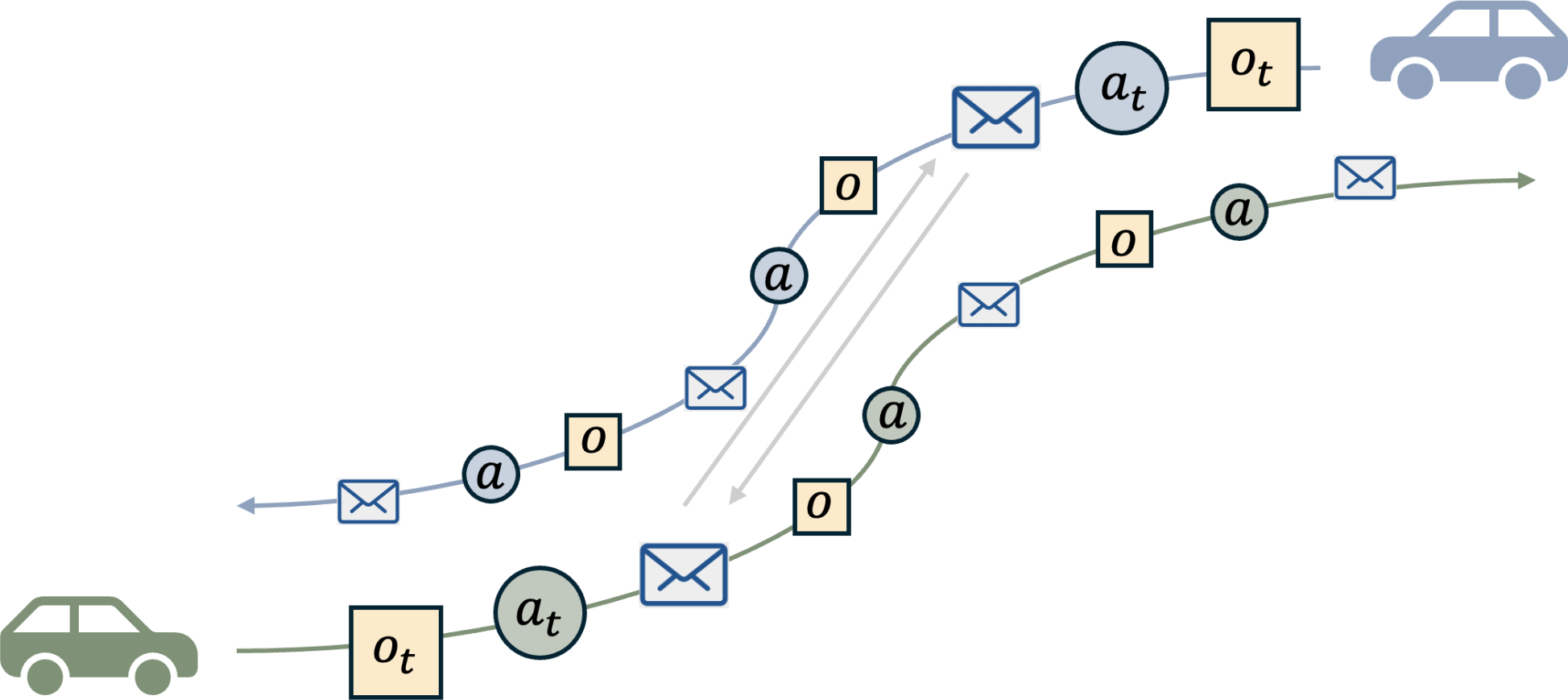}
    \caption{Overview of the In-Episode Communication Mechanism. At each time step, every agent simultaneously generates control actions and messages ($a_t$) conditioned on its observation ($o_t$). The exchanged messages are then passed among agents, and the resulting dialogue is incorporated into decision-making at the next time step.}
    \label{fig:interaction}
    \vspace{-12pt}
\end{figure}

The agent's policy is structured to output both control and communication commands. Specifically,  ${\pi_i(O_i, \{M_j\}_{j\in{\mathcal{F}}})\rightarrow \mathcal{A}_i}$ maps the observation of agent $i$ and the received messages $\{M_j\}_{j\in{\mathcal{F}}}$ to its action space ${\mathcal{A}_i=\langle \mathcal{M}_i, \mathcal{C}_i \rangle}$, where $\mathcal{M}_i$ represents the message generation space, which is constrained to natural language, and $\mathcal{C}_i$ denotes the vehicle control space with dimensions for throttle, brake, and steering inputs. At time step $t$, the message $M_i$ generated by agent $i$ is broadcast to all connected agents within a certain communication radius at the next time step $t+1$ (\figref{fig:interaction}).

This problem presents the following technical challenges:
\begin{enumerate}
    \item How can agents understand the situation and \textbf{generate} meaningful messages to collaboratively perceive the environment or negotiate in natural language;
    \item How can agents  \textbf{comprehend} incoming natural language messages and \textbf{incorporate} them into driving decision-making?
\end{enumerate}

\begin{figure}[ht]
    \centering
    \includegraphics[width=\linewidth]{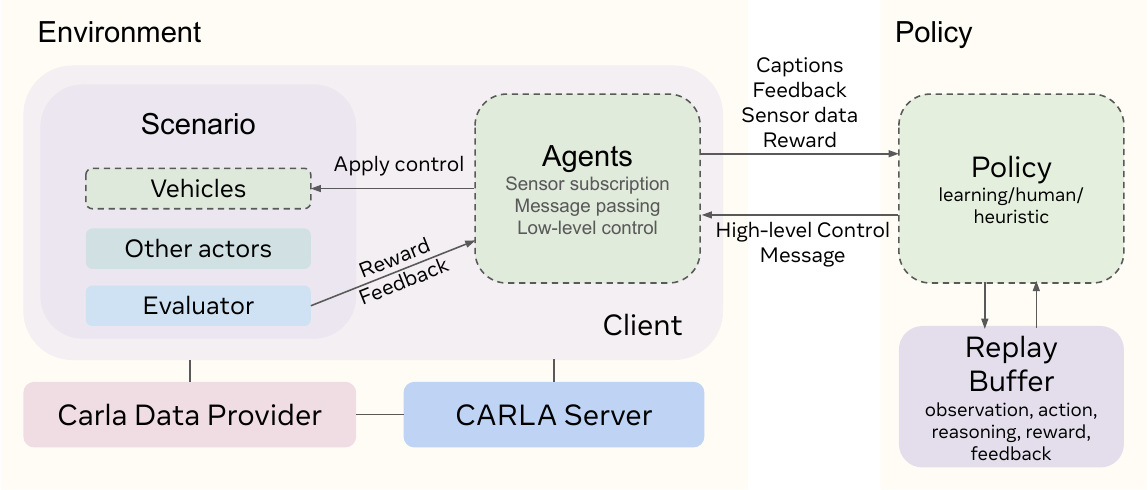}
    \caption{\small \textit{\simulator} Simulation framework. An agent is defined within the scenario and has a specific sensor registration and action space. A policy takes observations from an agent, computes actions, and learns from the experience replay buffer. }
    \label{fig:environment}
\end{figure}
\section{Environment}
\label{section:environment}
To provide concrete and typical driving scenarios that expose the \task~challenge, we have developed a simulation environment, \textbf{\simulator}, which is a multi-agent gymnasium environment \citep{brockman2016openaigym, terry2021pettingzoo} for the closed-loop evaluation of urban driving policies. \simulator~supports a flexible configuration of multi-agent scenarios, incorporating heterogeneous agents such as language agents, sensory agents, human agents, heuristic behavior agents, etc. It also enables \textbf{in-episode} communication between agents using a realistic simulated communication protocol based on \href{https://mqtt.org/}{\textsc{MQTT}}, a lightweight publish–subscribe protocol widely used in distributed systems. The simulation dynamics are implemented in CARLA \citep{dosovitskiy2017carla}, which provides realistic vehicle physics and complex urban traffic layouts.

\begin{figure*}[ht]
    \centering
    \includegraphics[width=\textwidth]{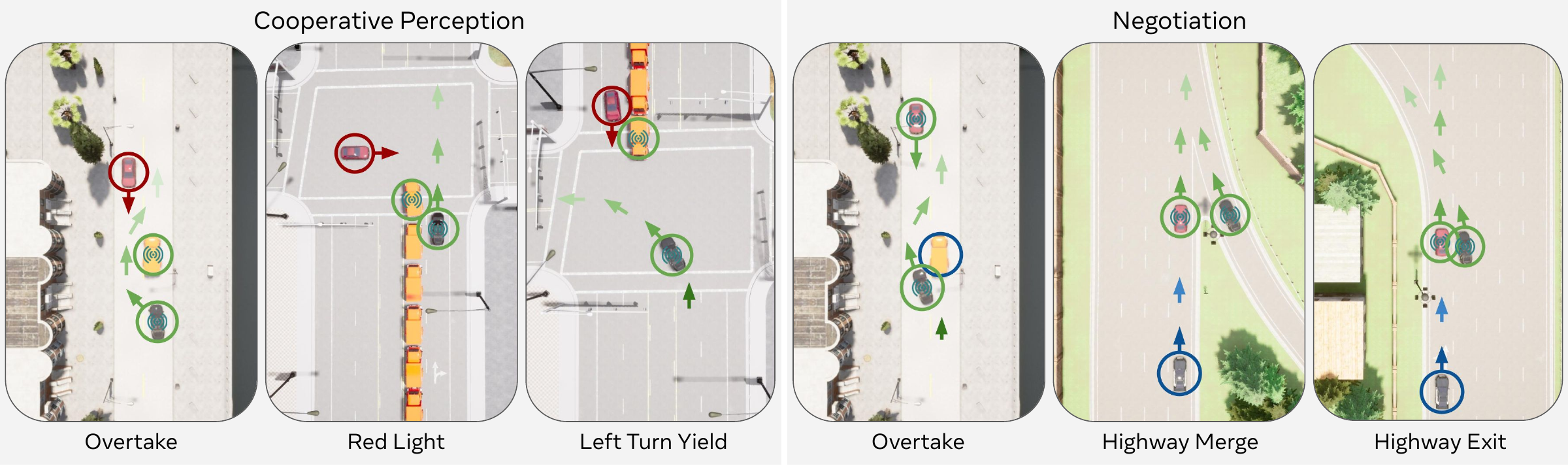}
    \caption{\small \emph{Overview of Scenarios and Agent Roles.} \textbf{\textcolor{ao(english)}{Green circles:}} Focal agents, agents aim at establishing coordination through communication; \textbf{\textcolor{brickred}{Red circles:} }Potential colliders; \textbf{\textcolor{mydarkblue}{Blue circles:}} Background agents.}
    \label{fig:scneario}
\end{figure*}
\subsection{Scenarios ($\mathcal{P}$) and Rewards ($\mathcal{R}$)} \simulator~has been set up with several accident-prone scenarios where multi-agent communication could be beneficial, as shown in \figref{fig:scneario}. 
Scenarios labeled with \textcolor{myblue}{{\texttt{{Cooperative Perception}}}} are cases where agents can benefit from receiving information about regions beyond their own line of sight, and scenarios labeled with \textcolor{myblue}{{\texttt{{Negotiation}}}} are cases where agents must communicate to resolve conflicts in their intended plans.
Each scenario features a focal group ($\mathcal{F}$) of agents operating alongside background agents with pre-scripted behaviors. Each focal agent is assigned a \textit{task} described in natural language, with success defined as reaching its target location within a time limit without collisions. 
Agents without motion targets, such as a stationary truck in cooperative perception tasks, do not earn rewards directly for themselves. However, the optimization objective encourages these agents to send messages that assist others to achieve their tasks. More scenario descriptions and reward setups are detailed in \appref{appendix: environment}.

\subsection{Observation Space ($\mathcal{O}$)} Our environment integrates a diverse range of sensor and simulator inputs inherited from CARLA. To focus on reasoning and multi-agent learning, we simplify environmental perception for \textbf{text-based agents} by introducing a rule-based, \textbf{partially observable captioner}. This module abstracts away the perception task, which would otherwise require object detection or vision-language models, by directly converting scenario information --- such as the states of the ego vehicle and others, lane details, and road conditions --- into natural language descriptions that convey \textit{factual} information while maintaining the partial observability imposed by the agent's line-of-sight sensors. For agents equipped with a transmitter/receiver device~(\textbf{transceiver}), real-time communication is enabled during episodes, and the message dialog is included as part of their observations. 
An example of a text-based observation is provided in \appref{appendix:prompts}.

\subsection{Action Space ($\mathcal{A}$)} The action space for each agent encompasses both vehicle control and message generation. The vehicle control space $\mathcal{C}$ is three-dimensional, consisting of throttle, brake, and steering. To reduce the decision-making frequency, agents execute high-level vehicle motion commands represented as temporal sequences of low-level vehicle controls $(C_t, C_{t+1}, ..., C_{t+k})$, where each command spans $k$ time steps. These high-level commands are atomic actions such as \texttt{go (adapt to a target speed)}, \texttt{stop}, \texttt{slow down}, \texttt{speed up}, and \texttt{change to the left lane}. They are composed through a combination of a global route planner and a local PID controller. The message generation space $\mathcal{M}$ is open-vocabulary, restricted to natural language tokens in this work, but the communication system is flexible enough to support other communication modes. 
In this work, messages are generated alongside the high-level control commands every 0.5 seconds ($k=10$ simulation steps). 
\section{Method}
\label{section:method}
\begin{figure*}[ht]
    \centering
    \includegraphics[width=\textwidth]{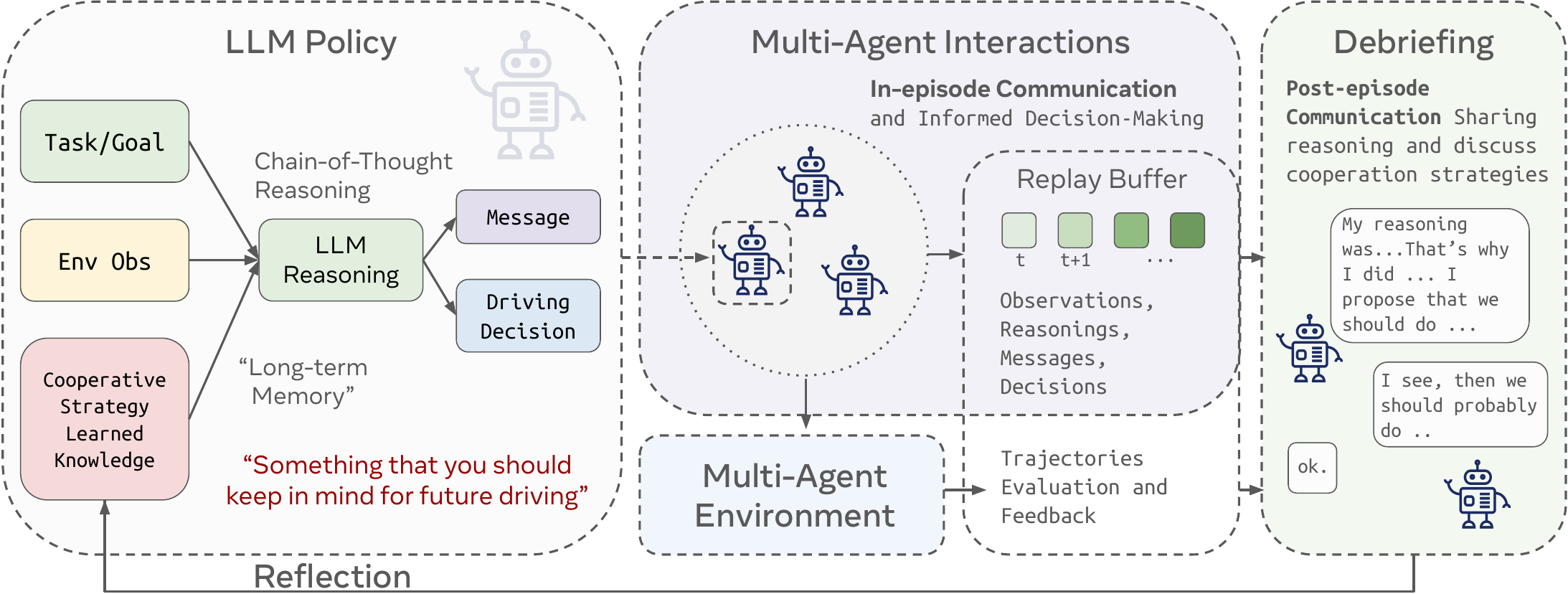}
    \caption{\small \textit{\method{} Agent Framework and Agent Learning Pipeline.}}
    \label{fig:method}
\end{figure*}

The core technical challenge of the \task{} problem is to enable embodied agents to {\textit{communicate in natural language for cooperative purposes and to adjust their actions dynamically according to the conversation}}. 
While prior works relied on extensive imitation learning data from human play to train agent policies that can speak or make decisions in natural language contexts \citep{meta2022cicero}, we instead explore leveraging pre-trained large language models to endow agents with such communication and reasoning capabilities.
To establish an initial solution, we adopt an \textbf{LLM agent framework} (\figref{fig:method}) that employs LLMs as a foundational prior for autonomous agents to engage in human-like communication, structuring messages within natural language space, and allowing agents to interpret messages to make informed driving decisions. 
However, LLMs are typically not trained for cooperative driving tasks. To address this limitation, we introduce \textbf{\method}, a \textbf{novel multi-agent learning method for LLM agents} built upon feedback loops that allow LLM agents to iteratively refine their communication and control policies through trial-and-error interactions with confederate agents. Inspired by how humans reflect and debrief after a cooperative game such as Hanabi, we enable agents to discuss cooperative strategies after each interaction episode. 

\subsection{Agent Policy}
Each agent $i$ follows a policy: $\pi_i(O_i, \{M_j\}_{j\in{\mathcal{F}}})\rightarrow \langle \mathcal{M}_i, \mathcal{C}_i \rangle$, where the distribution over actions follows the LLM used by the agent. Here, $O_i$ represents agent $i$'s comprehensive observation encompassing task and goal descriptions, environment details, and common traffic rules, expressed as a text or token sequence $\{t^{o}_{i}\}$. A received message $M_j = \{t^{m}_j\}$ from agent $j$ and a message to send $M_i = \{t^{m}_i\}$ are also text sequences generated by language agents. $C_i = \{t_i^c\}$ represents a text sequence for high-level control commands. The joint probability of selecting a command and generating a message is expressed as $P_i(\{t^{m}_i\}; \{t_i^c\} | \{t^{o}_{i}\}; \{{\{t^{m}_j\}\}}_{j\in\mathcal{F}})$ where \textbf{";"} indicates text concatenation and a large language model serves as the oracle to determine the probabilities.

\paragraph{\textbf{In-Context Knowledge}} Rather than fine-tuning LLM weights through gradient-based updates, \method{} adapts agent policies by modifying the \textit{context}, leveraging the auto-regressive nature of LLMs. Define \textcolor{myblue}{$\bm{K}_{\bm{i}} = \{{\bm{t}}_{\bm{i}}^{\bm{k}}\}$} as agent $i$'s accumulated knowledge and \textcolor{myblue}{${\bm{S}}_{\bm{i}} = \{{\bm{t}}_{\bm{i}}^{\bm{s}}\}$} as its cooperative strategy. The joint probability of generating commands and messages is then influenced by these additional prompt tokens:
$$P_i(\{t^{m}_i\}; \{t_i^c\} |  \textcolor{myblue}{\{{\bm{t}}_{\bm{i}}^{\bm{k}}\}}; \textcolor{myblue}{\{{\bm{t}}_{\bm{i}}^{\bm{s}}\}};\{t^{o}_{i}\}; \{{\{t^{m}_j\}\}}_{j\in\mathcal{F}}).$$

\paragraph{\textbf{Chain-of-Thought (CoT) Reasoning}}
LLM agents often generate better decisions by producing intermediate reasoning traces that structure their understanding of the situation~\citep{wei2022cot}. To leverage this observation, we prompt LLMs to reason step-by-step about the environment, incorporating observations, received messages, and in-context knowledge. The reasoning process generates an intermediate text sequence, denoted as \textcolor{myblue}{${\bm{R}}_{\bm{i}} = \{\bm{t}_{\bm{i}}^{\bm{r}}\}$}. The LLM agent then sample action token sequences $\{t_i^{m}\}$ and $\{t_i^{c}\}$ according to the probability distribution: $$P_i(\{t^{m}_i\}; \{t_i^c\} |  \textcolor{myblue}{\{t_i^k\}}; \textcolor{myblue}{\{t_i^s\}};\{t^{o}_{i}\}; \{{\{t^{m}_j\}\}}_{j\in\mathcal{F}}; \textcolor{myblue}{\{\bm{t}_{\bm{i}}^{\bm{r}}\}}).$$ The resulting output is formatted in JSON with two keys: \texttt{"command"} and \texttt{"message"}.

\subsection{\textbf{Agent Learning: Post-Episode Debriefing}}
The learning process is depicted in \figref{fig:method}. Initially, the LLM agents interact with each other in the scenarios and accumulate experience, which is stored in a replay buffer. Following the interaction phase, the agents engage in a debriefing session where they utilize past experiences as context to collaboratively refine a cooperative strategy. The outcomes of these discussions are summarized into two critical components: \textit{knowledge} (\textcolor{myblue}{$K_i = \{t_i^k\}$}) and \textit{cooperative strategies} (\textcolor{myblue}{$S_i = \{t_i^s\}$}). These components are subsequently integrated as in-context knowledge for future interactions, playing a pivotal role in shaping and improving the policy.

\paragraph{\textbf{Replay Buffer}} Each agent $i$ stores their trajectories locally. Each trajectory is a temporal sequence of transition data $T_i=\langle o_{i,t}, a_{i,t}, o_{i,t+1}\rangle$, which includes current and next local observations, commands, messages, and reasoning in a \textbf{replay buffer}, serving as a repository for further learning and iterative refinement. When an episode concludes, the environment evaluates each agent's performance and provides scalar rewards along with \textbf{verbal feedback}, such as \texttt{“Vehicle 109 collided with Vehicle 110 after 2 seconds.”} or \texttt{“Vehicle 111 stagnated for too long to complete its task.”} Each transition in the replay buffer is subsequently \textbf{retrospectively labeled} with enriched metadata, including responses from other agents, collision details (e.g., time to collision), stagnation specifics, and final rewards and outcomes. 

\paragraph{\textbf{Batch Context Sampling}}
Before engaging in the post-episode discussion (debriefing), each learning agent reflects on its individual experience from their perspective first.
While analyzing the entire trajectory would provide a comprehensive understanding of failure cases, computational and context window constraints necessitate sampling a subset (\textbf{batch}) of key frames from its replay buffer. To prioritize relevant data, the sampling process heuristically assigns (following \eqnref{eq: heuristic sampling}) higher probabilities to transitions that occur immediately before collisions, involve actions contributing to collisions, or lead to stagnation due to agents slowing down. Additionally, transitions that feature more intensive multi-agent interactions are given more weight. These selected samples serve as the context for subsequent analysis and strategy formulation, allowing the agent to focus on critical timesteps for improving performance. 

\paragraph{\textbf{Debriefing}}
When an episode ends due to the fault of a single agent, only that agent performs individual reflection. A debriefing (cooperative reflection) session is initiated when an episode ends in failure (either due to a collision or stagnation) arising from poor cooperation among agents. The debriefing session proceeds in a \textbf{turn-based} manner over $N$ rounds, with the aim of improving cooperative behavior in future interactions. The speaking order is deterministic in this work for each session, and agents take turns speaking in a round-robin format. The agent chosen to speak first is responsible for proposing a \textbf{joint} cooperative strategy (\textcolor{myblue}{$\bm{S}_1, \bm{S}_2,...\bm{S}_{i\in\mathcal{F}}$}) for everyone participating in the debriefing (the focal group). This agent begins by reasoning through its local transition data batch, analyzing the consequences of its actions, their influence on others and vice versa, and formulating a proposed strategy. Subsequently, the other agents take turns sharing their perspectives, providing feedback, or offering alternative insights based on their analysis of their own experience batch. After the discussion, each agent summarizes the discussion to develop \textbf{individual} cooperative strategies (\textcolor{myblue}{$\bm{S}_i$}) and knowledge (\textcolor{myblue}{$\bm{K}_i$}). These outcomes will later serve as in-context guidelines for future driving tasks. This joint discussion for future individual decision-making structure mirrors the principles of the Centralized Training Decentralized Execution (CTDE) framework~\citep{bernstein2002ctde}, a widely used approach in multi-agent learning. Our implementation details are available in \appref{appendix:implementation_details}.
\section{Experiments}
This section presents an empirical evaluation of \method{} and baseline approaches across different cooperative driving scenarios. We investigate the following research questions:
\begin{enumerate}
    \item Can LLM agents establish collaboration through chain-of-thought reasoning without prior interactions? \textcolor{myblue}{(The evaluated LLMs in this work can not.)}
    \item Does decentralized reflection enable LLM agents to improve their collaborative ability as they gain more interaction experiences? \textcolor{myblue}{(Yes.)}
    \item Does centralized discussion among LLM agents provide additional improvements in collaboration and communication compared to decentralized reflection? \textcolor{myblue}{(Yes.)}
    \item Can natural language communication enhance the performance and coordination of LLM agents compared to those without communication? \textcolor{myblue}{(Only if well trained.)}
\end{enumerate}

\noindent\paragraph{\textbf{Metrics}} Evaluation metrics are established based on the outcomes of agents who can incur reward (reward-eligible) for their tasks in the focal group, which is scenario-specific. For a scenario with N reward-eligible agents in the focal group, evaluated over M episodes, we utilize two key metrics:
\begin{enumerate}
    \item the \textbf{average collision rate} (\textbf{CR}), normalized by the group size, is
$\textcolor{myblue}{\frac{1}{N}\cdot\frac{1}{M}\sum_{m=1}^{M}\sum_{\textcolor{myblue}{i\in{\mathcal{F}}}} \mathbbm{1}\text{(agent }i\text{ involved in a collision)}}$, where collisions may involve both focal and background agents;
    \item the \textbf{average success rate} (\textbf{SR}), also normalized by the group size, is
$\textcolor{myblue}{\frac{1}{N}\cdot\frac{1}{M}\sum_{m=1}^{M}\sum_{\textcolor{myblue}{i\in{\mathcal{F}}}} \mathbbm{1}\text{(agent }i\text{ succeeded)}}$.
\end{enumerate}

Here, $\mathbbm{1}$ is the indicator function, equal to 1 if the event occurs and 0 otherwise.
The remaining failure cases, where agents exceed the time limit, heuristically determined to represent the upper bound for efficient task completion, without success or collision, are captured by the \textbf{average time out rate}, which can be derived as $\textcolor{myblue}{TR=1-SR-CR}$.

\paragraph{\textbf{Experimental Setup}}
For each baseline\footnote{Except for \method{}, which is only tested under the \textit{Comm} setting since it is particularly designed for improving multi-agent communication.}, we consider two settings labeled as \textit{Silent} and \textit{Comm}. In the \textit{\textbf{Silent}} setting,  LLM agents focus solely on controlling the vehicle based on their individual perception and reasoning without communication. 
The \textit{\textbf{Comm}} setting allows a method to generate either only messages or both messages and driving commands. For each LLM-based learning method, we allow agents to interact for up to 60 episodes per scenario, which is a random sequence alternating between safe (or randomized agent positions for highway negotiation settings) and accident-prone configurations with equal percentages.
We define a \textit{"solved"} criterion for learning success in a scenario as 20 consecutive successful episodes. Due to the uncontrollable randomness in the OpenAI models, we give each learning method 3 knowledge reset opportunities \footnote{Knowledge reset is done by clearing the learned knowledge before reaching a \emph{solved state indicator}, defined by 20 consecutive successful training episodes in this work.} to either report the \textit{"solved"} result, or otherwise, the last run for each seed. After learning, each method is evaluated for 30 episodes per scenario configuration per seed. We report experimental results aggregated with 3 seeds.

\paragraph{\textbf{Baselines}}
We established several baselines and scenarios to answer the research questions:
\begin{enumerate}
    \item \textbf{Zero-shot}: a base LLM agent using Chain-of-Thought (CoT) reasoning only, 
    \item \textbf{Reflection}: an LLM agent with CoT reasoning contextualized with knowledge from self-reflection, 
    \item \textbf{Correction+RAG (Silent)}: an LLM agent that corrects past actions via self-reflection, storing these corrections in a vector-based, retrievable memory, and uses few-shot retrieved example augmented generation (Correction+RAG). The retrieval augmented method without communication adapts DiLU \citep{wen2023dilu}, a non-communicating single-agent LLM-based approach that drives via reflection, to our environment. 
    \item \textbf{Correction+RAG (Comm)}: an LLM agent that resembles AgentsCoDriver \citep{hu2024agentscodriver}, the multi-agent communication extension of DiLU, but they do not actively optimize the messages.  
\end{enumerate}
For a fair comparison across baseline LLM agents, we do not initialize the knowledge with human data, nor is there human involvement during the learning process.
Moreover, we apply the same batch context sampling method for reflection or correction for all LLM agent baselines as our method. 
Additionally, we include \textbf{Coopernaut} \citep{cui2022coopernaut}, a LiDAR-based cooperative driving method, as an aspirational reference point for cooperative perception. Note that  Coopernaut is not directly comparable because it processes sensory data and communicates intermediate neural representations rather than natural languages. 
We do not compare with other multi-agent communication baselines for the same reasons.

\subsection{Quantitative and Qualitative Results}

\tabref{tab:experiment_quantitative} presents the quantitative evaluation of all methods across tasks. 
Notably, in this proof of concept, none of the LLM methods compared operate in real-time, requiring approximately 10 real-world seconds per decision step (0.5 seconds equivalent in simulation) using \textcolor{myblue}{\texttt{gpt-4o-mini}}. The inference latency primarily depends on reasoning, but we demonstrate an approach towards real-time inference in \secref{sec:generalization-and-distillation}.
%
On average, the natural language message bandwidth remains below 300 bytes per decision step, requiring less than 0.01 Mbps communication bandwidth. 
%
%
Based on these results, we provide responses to the research questions posed at the start of the section.
\begin{table*}[h!]
    \centering
    \caption{\small Experimental results for \textbf{per-scenario} learning. Each method is trained and evaluated independently in each scenario, using three random seeds and 30 evaluation episodes per seed. Results are reported as \texttt{mean $\pm$ standard deviation}. \textit{Zero-shot (Silent)} and \textit{Zero-shot (Comm)} denote LLM agents using chain-of-thought reasoning without or with communication, respectively. \textit{+Reflection} and \textit{+Correction+RAG} indicate single-agent learning through reflection or correction with retrieval-augmented generation. \textit{+Debrief} is our proposed multi-agent learning method \method{}. \textit{Coopernaut} is a non-LLM communication baseline for cooperative perception only.}
    \resizebox{\textwidth}{!}{
    \begin{tabular}{l|cc|cc|cc|cc|cc|cc}
    
    \toprule
    {\centering\multirow{5}{*}{\centering\diagbox[width=0.14\textwidth, height=0.1\textwidth]{\raisebox{1.2\height}{\normalsize Method}}{\raisebox{-1.5\height}{\normalsize Scenario}}}} &
    \multicolumn{6}{c|}{\multirow{2}{*}{\textcolor{myblue}{\normalsize \texttt{Cooperative Perception Scenarios}}}} & 
    \multicolumn{6}{c}{\multirow{2}{*}{\textcolor{myblue}{\normalsize\texttt{Negotiation Scenarios}}}} \\
    \multicolumn{1}{c|}{} &
    \multicolumn{6}{c|}{} &
    \multicolumn{6}{c}{}\\
    
    &
    \multicolumn{2}{c}{\multirow{2}{*}{\small \textbf{Overtake (Perception)}}} &
    \multicolumn{2}{c}{\multirow{2}{*}{\small \textbf{Red Light}}} &
    \multicolumn{2}{c|}{\multirow{2}{*}{\small \textbf{Left Turn}}} &
    \multicolumn{2}{c}{\multirow{2}{*}{\small \textbf{Overtake (Negotiation)}}} &
    \multicolumn{2}{c}{\multirow{2}{*}{\small \textbf{Highway Merge}}} &
    \multicolumn{2}{c}{\multirow{2}{*}{\small \textbf{Highway Exit}}} \\

    \multicolumn{1}{c|}{} &
    \multicolumn{6}{c|}{} &
    \multicolumn{6}{c}{}\\

    \cmidrule{2-13}
    & 
    CR (\%) $\downarrow$ & SR (\%) $\uparrow$ & 
    CR (\%) $\downarrow$ & SR (\%) $\uparrow$ & 
    CR (\%) $\downarrow$ & SR (\%) $\uparrow$ &
    CR (\%) $\downarrow$ & SR (\%) $\uparrow$ & 
    CR (\%) $\downarrow$ & SR (\%) $\uparrow$ & 
    CR (\%) $\downarrow$ & SR (\%) $\uparrow$ \\
    
    \midrule
    Zero-shot (Silent)  &
        93.3 $\pm$ 3.4 & 0.0 $\pm$ 0.0 &                            
        93.3 $\pm$ 6.7 & 6.7 $\pm$ 6.7 &                            
        93.3 $\pm$ 5.8 & 6.7 $\pm$ 5.8 &                           
        89.9 $\pm$ 2.8 & 7.2 $\pm$ 3.8 &     
        100.0 $\pm$ 0.0 & 0.0 $\pm$ 0.0 &    
        33.3 $\pm$ 9.3 & 66.1 $\pm$ 9.2\\    
        
    +Reflection &
        87.8 $\pm$ 3.4 & 0.0 $\pm$ 0.0 &                           
        94.4 $\pm$ 6.9 & 5.6 $\pm$ 6.9 &                            
        76.7 $\pm$ 20.8 & 23.3 $\pm$ 20.8 &                          
        32.8 $\pm$ 29.4 & 36.7 $\pm$ 52.1 &  
        15.0 $\pm$ 23.1 & 84.4 $\pm$ 22.6 &  
        32.8 $\pm$ 13.4 & 67.2 $\pm$ 13.4 \\ 
        
    +Correction+RAG &
        62.0 $\pm$ 31.9 & 4.4 $\pm$ 7.7 &                            
        93.3 $\pm$ 3.3 & 6.7 $\pm$ 3.3 &                           
        64.4 $\pm$ 15.0 & 35.6 $\pm$ 15.0 & 
        46.7 $\pm$ 21.9 & 33.3 $\pm$ 28.0 &  
        35.6 $\pm$ 29.4 & 64.4 $\pm$ 29.4 &  
        33.9 $\pm$ 28.4 & 51.1 $\pm$ 14.2 \\ 
        
    \midrule
    Zero-shot (Comm) &
        91.1 $\pm$ 5.1 & 4.4 $\pm$ 5.1 &                           
        60.0 $\pm$ 11.5 & 38.9 $\pm$ 10.7 &                           
        85.6 $\pm$ 8.4 & 14.4 $\pm$ 8.4 &                          
        87.8 $\pm$ 5.9 & 11.7 $\pm$ 6.7 &   
        67.2 $\pm$ 27.1 & 32.8 $\pm$ 27.1 & 
        53.3 $\pm$ 11.5 & 46.7 $\pm$ 11.5 \\

    +Reflection &
        63.3 $\pm$ 14.5 & 34.4 $\pm$ 10.7 &                            
        37.8 $\pm$ 18.4 & 47.8 $\pm$ 18.4 &                            
        51.1 $\pm$ 37.2 & 47.8 $\pm$ 36.0 &
        55.6 $\pm$ 38.9 & 43.3 $\pm$ 37.1 & 
        20.0 $\pm$ 1.7 & 80.0 $\pm$ 1.7 &   
        53.9 $\pm$ 24.1 & 45.6 $\pm$ 23.6 \\
        
    +Correction+RAG &
        4.4 $\pm$ 1.9 & 90.0 $\pm$ 6.7 &           
        13.3 $\pm$ 12.0 & 66.7 $\pm$ 27.3 &                            
        43.3 $\pm$ 38.4 & 38.9 $\pm$ 22.7 &                          
        38.3 $\pm$ 6.0 & 61.1 $\pm$ 5.4 &   
        40.0 $\pm$ 18.0 & 60.0 $\pm$ 18.0 & 
        49.4 $\pm$ 49.2 & 43.3 $\pm$ 39.8 \\
        
    +Debrief (ours) &
        \textbf{1.1 $\pm$ 1.9} & \textbf{94.4 $\pm$ 6.9} &                   
        \textbf{0.0 $\pm$ 0.0} & \textbf{93.3 $\pm$ 5.8} & 
        \textbf{6.7 $\pm$ 3.3} & \textbf{92.2 $\pm$ 3.8} & 
        \textbf{3.3 $\pm$ 3.3} & \textbf{95.6 $\pm$ 3.8} & 
        \textbf{6.7 $\pm$ 11.5} & \textbf{93.3 $\pm$ 11.5} & 
        \textbf{18.3 $\pm$ 21.7} & \textbf{81.1 $\pm$ 21.2} \\ 
        
    \midrule
    
    \textcolor{gray}{Coopernaut (Comm)} &
        \textcolor{gray}{4.5 $\pm$ 3.1} & \textcolor{gray}{90.5 $\pm$ 1.2} & 
        \textcolor{gray}{17.7 $\pm$ 7.8} & \textcolor{gray}{80.7 $\pm$ 7.6} & 
        \textcolor{gray}{18.1 $\pm$ 6.2} & \textcolor{gray}{80.7 $\pm$ 5.2} &
        \textcolor{gray}{N/A} & \textcolor{gray}{N/A} &
        \textcolor{gray}{N/A} & \textcolor{gray}{N/A} &
        \textcolor{gray}{N/A} & \textcolor{gray}{N/A} \\
    \bottomrule
    \end{tabular}
    }
    \label{tab:experiment_quantitative}
\end{table*} 

\paragraph{\textbf{R1: LLM agents with CoT examined in this paper do not establish collaboration through communication in zero-shot interactions.}}
Our experiments show that Zero-Shot agents (\texttt{gpt-4o-mini}), even with communication enabled, fail to coordinate effectively. The failure modes are (1) agents do not communicate effectively to understand each other's needs in perception or achieve agreement in negotiation, or (2) even when the messages make sense to humans, agents do not respond with appropriate driving commands. This result suggests that without prior training or explicit strategies, chain-of-thought reasoning alone is insufficient to foster effective coordination. At the time of writing, preliminary experiments with other LLMs such as \texttt{Llama 3} and \texttt{gpt-4o} follow a similar pattern. 
Future work could systematically examine whether large reasoning models like \texttt{gpt-o4} can mitigate these limitations.

\paragraph{\textbf{R2: Decentralized learning can enable LLM agents to improve their collaborative ability as they gain more interaction experiences. }}
The decentralized learning methods, Reflection and Correction+RAG, show significant improvement in reducing collision rates from Zero-Shot across tasks. Reflection allows agents to individually analyze their experience to generate knowledge, but the knowledge is often more reactive than proactive (see \appref{appendix:highway_merge_silent_tip_gpt} for example).
The Correction+RAG method records successful episodes to preserve successful coordination patterns and correct commands and messages at key frames selected through a heuristic batch sampling.
However, although the method improves the control response strategy, we find that it qualitatively does not always produce messages that are consistent with the actions, likely due to its open-loop message revision process.
Both methods show promise, but have room for improvement.

\paragraph{\textbf{R3: Centralized debriefing enhances coordination more than decentralized reflection.}}
The debriefing method, which focuses on generating explicit cooperation strategies, enables LLM agents to achieve more stable collaboration compared to decentralized reflection or zero-shot approaches, evidenced by higher success rates than baselines across tasks.
Qualitatively, the resulting conversations are human-interpretable, paving the way for future human–AI communication in cooperative driving scenarios. Please refer to our \underline{supplementary videos} for examples of the generated dialogues. 
The primary performance boost of \method{} stems from the formalized coordination strategy, which explicitly defines how each agent should communicate and respond within a dialogue across different scenarios. Examples of learned cooperative strategies for each scenario are illustrated in \appref{appendix:qualitative-knowledge}. 
Interestingly, \method{} reveals that LLMs can struggle to understand complex AI-generated messages that resemble natural language, so agents eventually develop concise communication protocols (like "hold" and "go" in \appref{sec:communication-protocol}) to ensure that their intentions are easily interpretable among themselves. 
However, open challenges still remain. For example, although the learned strategies are easy to verify, the debriefing process used to identify them can sometimes fail, where no agents can find issues with the cooperation strategies in harder and longer-horizon tasks, such as \texttt{negotiation-highway-exit}.

\paragraph{\textbf{R4: Natural language communication in cooperative driving can be effective,  but may pose safety risks without good communication strategies.} }
Our method, which operates with natural language communication, provides a proof of concept for natural-language-based multi-agent coordination across scenarios. However, learning to communicate effectively remains challenging. In cooperative perception tasks, communication-enabled methods consistently outperform silent ones, highlighting the critical role of information sharing. 
In contrast, in negotiation scenarios such as \texttt{highway-merge} and \texttt{highway-exit}, agents generally perform better in silent mode. This result suggests that communication can add complexity and hinder coordination when not well-optimized. We speculate that the root cause lies in the suboptimal communication strategies learned under decentralized training, where messages may introduce noise rather than useful signals.


\begin{figure}[h!]
    \centering
    \begin{subfloat}[Distillation.]{%
        \includegraphics[width=0.33\textwidth]{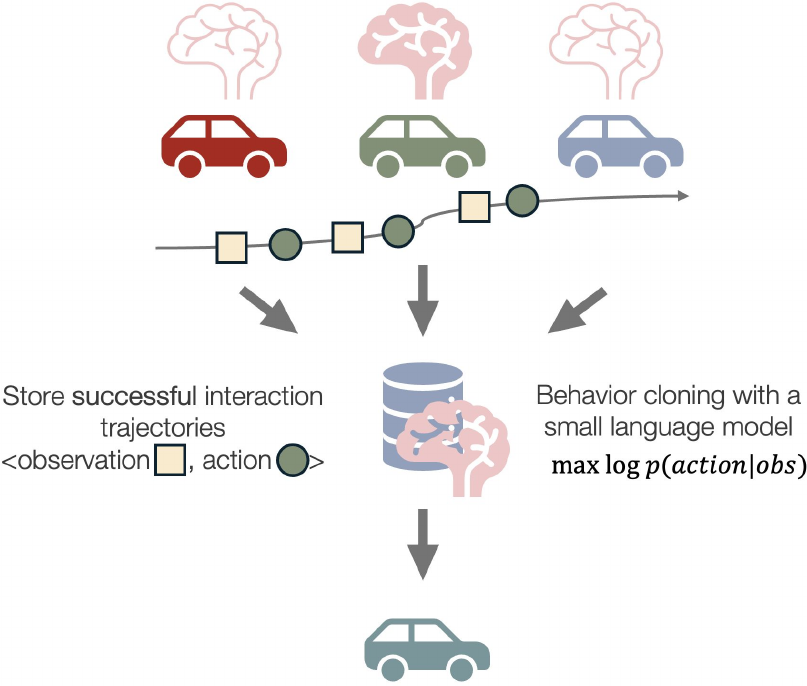}
        \label{fig:distillation}
    }
    \end{subfloat}
    \begin{subfloat}[Centralized Memory.]{%
        \includegraphics[width=0.22\textwidth]{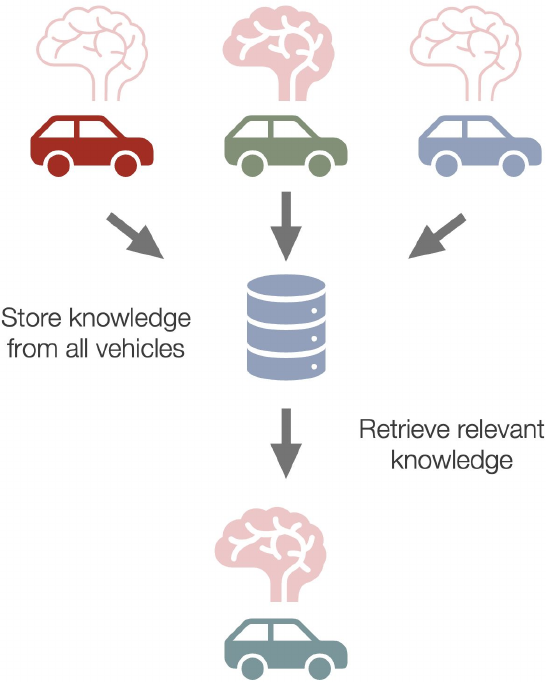}
        \label{fig:centralized_memory}
    }
    \end{subfloat}
    \caption{\small Generalization Methods. \textcolor{myblue}{\textbf{(a)}} Distillation performs full-parameter fine-tuning by matching the probability distribution of expert decisions from successful trajectories, producing a compact model that acts directly on observations without explicit reasoning. \textcolor{myblue}{\textbf{(b)}} Centralized Memory aggregates learned knowledge and cooperative strategies from all vehicles across all scenarios into a shared memory. Each vehicle accesses this memory to retrieve the most relevant knowledge based on its observation and dialogue, followed by Chain-of-Thought reasoning and decision-making.}
    \label{fig:generalization}
    \vspace{-24pt}
\end{figure}
\subsection{Towards Real-Time Cross-Scenario Cross-Role Generalization}
\label{sec:generalization-and-distillation}

Up to this point, a separate policy was trained to handle each of the \simulator{} scenarios. However, for practical deployment, it is desirable to develop a single policy that can handle a broad range of challenging driving scenarios. We explore two independent approaches for achieving cross-scenario generalization: \textit{\textbf{Centralized Memory}} and \textit{\textbf{Distillation}} (\figref{fig:generalization}).  \textit{\textbf{Centralized Memory}} aggregates all agents' \textit{most effective} knowledge---identified by the highest estimated success rate across learning trials---into a unified vector memory. Agents then search in the memory according to the observation and dialogue for the most relevant knowledge.  
\textit{\textbf{Distillation}} performs full-parameter fine-tuning of a small language model, DistilGPT2 \citep{radford2019gpt2, sanh2019distilgpt2}, to directly \textbf{imitate} the behavior of the \method{} agent with the most effective knowledge. The imitation dataset is aggregated from all \textit{successful} evaluation episodes across scenarios, and the distillation model is trained to minimize the token-level cross-entropy loss against the large model’s outputs. During inference, decisions are generated via random sampling with a temperature of 0.2. 

As shown in \tabref{tab:distilled_latencies}, the Distillation model achieves decision generation times between \textbf{100 ms} and \textbf{470 ms} on an NVIDIA A40 GPU, depending on message generation length (\textbf{50 bytes} to \textbf{300 bytes}), getting close to the 500 ms decision-making frequency, though time delays and asynchrony have not been fully considered. Evaluation results of the two methods in accident-prone scenarios are listed in \tabref{tab:experiment_generalization}. Remarkably, the distilled model generalizes well across scenarios and even surpasses the performance of its teacher model (\method{}) in some cases. However, we observe that it tends to behave overly conservatively in safe \texttt{perception-overtake} scenarios (\tabref{tab:additional_cooperative_perception}, \tabref{tab:additional_negotiation}), suggesting room for further improvement, potentially through expert-guided correction methods such as DAgger \citep{ross2011dagger}.

\begin{table*}[ht]
    \centering
    \caption{\small Cross-scenario generalization results. Policies are trained once and evaluated over 3 seeds (30 episodes per seed). 
    We report the mean $\pm$ standard error across seeds. 
    \textit{Debrief (per-scenario)} is an oracle baseline trained separately for each scenario, used to benchmark the generalization of \textit{Centralized Memory} and \textit{Distillation}.}
    \vspace{-8pt}
    \resizebox{\textwidth}{!}{
    \begin{tabular}{l|cc|cc|cc|cc|cc|cc}
    \toprule
    {\centering\multirow{5}{*}{\centering\diagbox[width=0.14\textwidth, height=0.1\textwidth]{\raisebox{1.2\height}{\normalsize Method}}{\raisebox{-1.5\height}{\normalsize Scenario}}}} &
    \multicolumn{6}{c|}{\multirow{2}{*}{\textcolor{myblue}{\normalsize \texttt{Cooperative Perception Scenarios}}}} & 
    \multicolumn{6}{c}{\multirow{2}{*}{\textcolor{myblue}{\normalsize\texttt{Negotiation Scenarios}}}} \\
    \multicolumn{1}{c|}{} &
    \multicolumn{6}{c|}{} &
    \multicolumn{6}{c}{}\\
    
    &
    \multicolumn{2}{c}{\multirow{2}{*}{\small \textbf{Overtake (Perception)}}} &
    \multicolumn{2}{c}{\multirow{2}{*}{\small \textbf{Red Light}}} &
    \multicolumn{2}{c|}{\multirow{2}{*}{\small \textbf{Left Turn}}} &
    \multicolumn{2}{c}{\multirow{2}{*}{\small \textbf{Overtake (Negotiation)}}} &
    \multicolumn{2}{c}{\multirow{2}{*}{\small \textbf{Highway Merge}}} &
    \multicolumn{2}{c}{\multirow{2}{*}{\small \textbf{Highway Exit}}} \\

    \multicolumn{1}{c|}{} &
    \multicolumn{6}{c|}{} &
    \multicolumn{6}{c}{}\\

    \cmidrule{2-13}
    & 
    CR (\%) $\downarrow$ & SR (\%) $\uparrow$ & 
    CR (\%) $\downarrow$ & SR (\%) $\uparrow$ & 
    CR (\%) $\downarrow$ & SR (\%) $\uparrow$ &
    CR (\%) $\downarrow$ & SR (\%) $\uparrow$ & 
    CR (\%) $\downarrow$ & SR (\%) $\uparrow$ & 
    CR (\%) $\downarrow$ & SR (\%) $\uparrow$ \\
    
    \midrule
    \textcolor{gray}{Debrief (per-scenario)} &
        \textcolor{gray}{1.1 $\pm$ 1.1} & \textcolor{gray}{\textbf{98.9 $\pm$ 1.1}} & 
        \textcolor{gray}{0.0 $\pm$ 0.0} & \textcolor{gray}{96.7 $\pm$ 0.0} & 
        \textcolor{gray}{4.4 $\pm$ 2.9} & \textcolor{gray}{94.4 $\pm$ 2.2} & 
        \textcolor{gray}{10.0 $\pm$ 3.8} & \textcolor{gray}{87.2 $\pm$ 3.9} & 
        \textcolor{gray}{2.2 $\pm$ 2.2} & \textcolor{gray}{97.8 $\pm$ 2.2} & 
        \textcolor{gray}{13.3 $\pm$ 6.0} & \textcolor{gray}{86.7 $\pm$ 6.0} \\ 
    Centralized Memory &
        2.2 $\pm$ 1.1 & 93.3 $\pm$ 1.9 &  
        0.0 $\pm$ 0.0 & \textbf{100.0 $\pm$ 0.0} &  
        4.4 $\pm$ 2.9 & 93.3 $\pm$ 3.3 &  
        12.2 $\pm$ 2.9 & 86.7 $\pm$ 1.9 &  
        1.1 $\pm$ 1.1 & 98.9 $\pm$ 1.1 &  
        16.1 $\pm$ 4.8 & 82.8 $\pm$ 5.3 \\ 
    Distillation &
        \textbf{0.0 $\pm$ 0.0} & 83.3 $\pm$ 1.9 &
        0.0 $\pm$ 0.0 & 91.1 $\pm$ 4.4 &
        \textbf{0.0 $\pm$ 0.0} & \textbf{96.7 $\pm$ 0.0} &
        \textbf{10.0 $\pm$ 3.3} & \textbf{88.9 $\pm$ 4.4} &
        \textbf{0.0 $\pm$ 0.0} & \textbf{100.0 $\pm$ 0.0} &
        \textbf{3.3 $\pm$ 0.0} & \textbf{96.7 $\pm$ 0.0}\\
    \bottomrule
    \end{tabular}
    }
    \label{tab:experiment_generalization}
\end{table*}

\begin{table*}[h]
    \centering
    \caption{\small Decision Latency, Message Size using \texttt{Distilled} LLM Policy}
    \vspace{-8pt}
    \resizebox{0.75\textwidth}{!}{
    \begin{tabular}{l|ccc|ccc}
    \toprule
    {\diagbox{Latencies}{Scenario}} & \textbf{Overtake (Perception)} & \textbf{Left Turn} & \textbf{Red Light} & \textbf{Overtake (Negotiation)} & \textbf{Highway Merge} & \textbf{Highway Exit}\\
    \midrule
     Decision Latency (s) & 0.45 & 0.44 & 0.38 & 0.14 & 0.19 & 0.20 \\
     Message Size (bytes) & 223.3 & 297.9 & 223.0 & 28.0 & 59.0 & 59.0 \\
    \bottomrule
    \end{tabular}
    }
    \label{tab:distilled_latencies}
\end{table*} 
\section{Related Work}

\paragraph{\textbf{LLM Agents for Autonomous Driving.}} have shown potential to address various autonomous driving tasks. In particular, they are promising in tackling corner cases~\citep{wen2023road} due to their reasoning ability and the common-sense knowledge embedded, yielding a more generalizable autonomous driving stack. 
Recent studies have explored various approaches to tailor state-of-the-art LLMs for driving \citep{wen2023dilu, hu2024agentscodriver}.
However, a foundational challenge lies in grounding LLM agents in the real world---they need to perceive and understand the traffic scenarios. 
A straightforward approach is to obtain the observations from oracle perception models~\citep{mao2023agentdriver} and convert them to textual descriptions~\citep{mao2023gptdriver,sha2023languagempc,jin2023surrealdriver,cui2023receive}.
Some other studies tackled this challenge by introducing Visual Language Models (VLMs), which are adapted to driving domains through in-context instruction tuning~\citep{ma2023dolphins} or fine-tuning~\citep{wayve2023lingo1,xu2023drivegpt4,ding2023hilm,yang2023lidar}.
To enhance LLM agents' reasoning ability, prior works have investigated incorporating handcrafted guidance and examples in the prompts~\citep{sha2023languagempc,jin2023surrealdriver,cui2023receive}, structuring the reasoning procedure~\citep{mao2023agentdriver,sima2023drivelm}, and fine-tuning the models on driving datasets. 
Notably, fine-tuning LLMs and VLMs requires an extensive amount of driving data with language labels. 
Several works have attempted to adapt existing language-driving datasets for LLM fine-tuning~\citep{ding2023hilm,xu2023drivegpt4,ma2023dolphins} or augment large-scale multimodal driving datasets \citep{caesar2020nuscenes, sun2020scalability, mao2021one} with language labels~\citep{qian2023nuscenesqa, shao2023lmdrive, sima2023drivelm, nie2023reason2drive}. In contrast, our work generates scalable driving data through agent self-play. 
Note that existing models were predominantly evaluated in an \emph{open-loop} fashion.
In contrast, similar to some prior works ~\citep{shao2023lmdrive,sha2023languagempc,jin2023surrealdriver}, we conduct closed-loop evaluation of the proposed method and baseline methods in CARLA~\citep{dosovitskiy2017carla}.

\vspace{-6pt}
\paragraph{\textbf{Natural Language Communication for Driving.}}
There is a scarcity of prior research on optimizing LLM agents in multi-agent settings with natural language vehicle-to-vehicle communication, with only a few concurrent but distinct efforts such as \citep{hu2024agentscodriver, gao2025langcoop}.
AgentsCoDriver \citep{hu2024agentscodriver} leverages a vector memory that stores and corrects vehicle control actions associated with specific situations, but it only optimizes the control actions through self-reflection while leaving message generation to the emergent capability of LLMs.
Other studies on multi-agent collaboration for autonomous robots do not employ trial-and-error optimization; instead, they rely on human-engineered structures or prompting schemes to guide coordination. For example, LangCoop \citep{gao2025langcoop} designs a structured in-context knowledge format to facilitate intent inference among interactive agents, CoMAL \citep{yao2025comal} adds a collaboration module prompting agents to determine their respective roles, and GameChat \citep{mahadevan2025gamechat} constructs a multi-round communication process to ensure agents reach consensus in constrained navigation tasks. In summary, \simulator{} serves as a comprehensive testbed for studying multi-agent natural language communication among autonomous vehicles, and \method{} represents the first CTDE-style multi-agent reinforcement learning approach for multi-LLM-agent systems that jointly optimizes both communication and control in a closed-loop manner.
\section{Conclusion and Future Work}
This work explores how autonomous vehicles can communicate and coordinate through natural language, offering a path toward future human–AI collaboration in cooperative driving. We introduce \simulator{}, a multi-agent simulation environment for closed-loop evaluation of collaboration through vehicle-to-vehicle dialogue, and propose \method{}, a multi-agent learning framework that enables LLM-based driving agents to refine communication and control through iterative reflection and debriefing. Experiments show that while zero-shot LLM agents fail to collaborate effectively, reflective and centralized learning yields stable, human-interpretable cooperation across both perception and negotiation tasks. By distilling the learned behaviors into a compact model, we further achieve near-real-time, cross-scenario generalization. Overall, this study establishes natural language as a promising medium for cooperative autonomy and highlights future works for grounding communication in more realistic perception and communication mechanisms, improving multi-agent learning stability, and evaluating the agents in real human–AI collaborative driving.





\begin{acks}
    {\small This work has taken place in the Learning Agents Research
Group (LARG) at UT Austin.  LARG research is supported in part by NSF
(FAIN-2019844, NRT-2125858), ONR (N00014-24-1-2550), ARO
(W911NF-17-2-0181, W911NF-23-2-0004, W911NF-25-1-0065), DARPA (Cooperative Agreement HR00112520004 on Ad Hoc Teamwork) Lockheed
Martin, and UT Austin's Good Systems grand challenge.  Peter Stone
serves as the Chief Scientist of Sony AI and receives financial
compensation for that role.  The terms of this arrangement have been
reviewed and approved by the University of Texas at Austin in
accordance with its policy on objectivity in research.}
\end{acks}




\bibliographystyle{ACM-Reference-Format} 
\bibliography{reference}


\appendix
\section{Method}
\label{appendix:method}
The \algoref{alg:coopreflect} implements \method. This multi-LLM-agent learning framework leverages communication and centralized cooperative reflection using large language models (LLMs) to enhance coordination between agents in a simulated environment. 

\begin{algorithm}
    \caption{\method}
    \label{alg:coopreflect}
    \begin{algorithmic}
        \STATE \textbf{Input:} Multi-agent Simulation Environment \texttt{env}, LLM agents\{$\pi_{i\in \mathcal{I}}$\}, Debriefing round $R$.
        \STATE \textbf{Initialize:} Knowledge \{$K_{i\in \mathcal{I}}$\}, Cooperative Strategy \{$S_{i\in \mathcal{I}}$\}, Replay Buffer \texttt{replay\_buffer}

        \FOR{$j{=}1,2,3...$ \textcolor{gray}{// Training epoch or early stop criteria met}}
                \STATE \{obs$_i$\} = \texttt{env.reset}()\\
                \WHILE{$t<T$ \textcolor{gray}{// Time step}}
                    \FOR{$i=1,..., N$ \textcolor{gray}{//Per agent, but execute in parallel}}
                        \STATE \textcolor{gray}{// Agent reasons the observation}
                        \STATE {reasoning$_i$ $\gets$ \texttt{agent}$_i$.\texttt{reason}(obs$_i$, K$_i$, S$_i$)}
                        \STATE \textcolor{gray}{// Agent makes decision based on reasoning}\\
                        \STATE message$_i$, control$_i$ $\gets$ \texttt{agent}$_i$.\texttt{act}(obs$_i$, reasoning$_i$)
                    \ENDFOR\\
                    \STATE \textcolor{gray}{// Step the environment with joint actions}
                    \STATE \{next$\_$obs$_i$\} $\gets$ \texttt{env.step}(\{message$_i$, control$_i$\})
                    
                    \STATE \textcolor{gray}{// Message Dialog becomes part of the observation}
                    \STATE \{obs$_i$\} $\gets$ \{next$\_$obs$_i$\} $\cup$ \{message$_i$\}
                    \FOR{$i=1,..., N$ \textcolor{gray}{//Per agent, but execute in parallel}}
                    \STATE \textcolor{gray}{// Store experience to the replay buffer}
                    \STATE \texttt{agent}$_i$.\texttt{replay\_buffer}\texttt{.add}(obs$_i$, message$_i$, control$_i$, next\_obs$_i$, reasoning$_i$, additional information)
                    \ENDFOR\\
                \ENDWHILE\\
                \STATE \textcolor{gray}{// Get episode feedback from the environment}
                \STATE feedback $\gets$ \texttt{env.evaluate}() \\
                \IF{feedback.success = True}
                \STATE continue \textcolor{gray}{// Skip debriefing when the episode is successful}
                \ENDIF
                \STATE \textcolor{gray}{// Lable all the transition data in hindsight}
                \STATE \texttt{data\_post\_processing(\texttt{all\_agents}.\texttt{replay\_buffer})}
                \STATE \textcolor{blue}{// Debriefing and learning from feedback, update knowledge}
                \STATE \textcolor{gray}{// Randomly decide debrief order}
                \FOR{$r=1,..., N$}
                \STATE\texttt{agent$_r$.analyze(agent$_r$.sample(batch))}
                    \IF{strategy=None}
                    \STATE coop\_stategy = \texttt{agent$_r$.propose()}
                    \ELSE
                    \STATE coop\_stategy = \texttt{agent$_r$.revise(}coop\_stategy\texttt{)}
                    \ENDIF
                \ENDFOR
                \STATE \textcolor{gray}{//Summarize the dialogue and use it for future learning}
                \STATE $\{K_i, S_i\}$ $\gets$ \texttt{agent.reflect(coop\_stategy)}
        \ENDFOR\\
        last $\{K_i, S_i\}$ during the last iteration
    \end{algorithmic}
\end{algorithm}

\subsection{Implementation Details}
\label{appendix:implementation_details}
We utilize \textcolor{myblue}{\texttt{gpt-4o-mini}} with a temperature of 0.2 for the agent policy, making decisions and collecting experiences every 0.5 seconds in simulation (10 simulation frames). The message dialogs received are maintained within a 2-second window according to the age of the message during each episode. The debriefing process is conducted after each episode for a total of 60 episodes, comprising a $N=1$ round of strategy proposal and revision among agents followed by a final round of individual reflection to summarize and consolidate the discussion results. To enable stronger reasoning and summarization capabilities, \textcolor{myblue}{\texttt{gpt-4o}} is used for the debriefing sessions and reflection. The transition data for trajectory analysis are sampled from the trajectory with a batch size of 4. 

The \textbf{Batch Context Sampling} follows the following heuristic rule to assign probability mass on each transition data point and sample according to the normalized probability mass:
\begin{equation}
\begin{split}
    \text{Weight}_i = 1 
    &+ 2\times\mathbbm{1}\{\text{exists other agents}\} \\
    &+ 5 \times \max(2 - \text{time to collision}, 0) \\
    &+ 10 \times \mathbbm{1}\{\text{actions contribute to collision}\} \\
    &+ 0.1 \times \mathbbm{1}\{\text{stagnation}\}\times \{\text{timestep}\}\\
    &+ 2 \times \mathbbm{1}\{\text{actions contribute to stagnation}\}
\end{split}
\label{eq: heuristic sampling}
\end{equation}
where $\mathbbm{1}$ represents the indicator function that takes the value 1 when the event happens, and 0 otherwise.

\textbf{Note on Knowledge Reset:} Due to the inherent stochasticity in OpenAI models during the time of our experiments, the knowledge acquired by the LLM agents may become unpredictably corrupted throughout training. Therefore, we allow each LLM agent learning method up to three \emph{knowledge resets} (clearing the knowledge) before reaching a \emph{solved state indicator} (defined by 20 consecutive successful episodes) or using the final attempt after the last reset. This strategy resembles best-of-N sampling; however, large-scale evaluation of learned knowledge is costly, so our knowledge selection relies on an automated heuristic indicator during training.

\subsection{Inference Latencies}
\label{appendix:latencies}
\tabref{tab:latencies} summarizes the average latencies and message sizes for each scenario in the communication setting, evaluated using \texttt{gpt-4o-mini} on a machine with Intel Gen 10 CPUs. The metrics include partial observable captioner latency (in seconds), reasoning latency (in seconds), decision latency (in seconds, excluding reasoning latency), and message size (in Mb). Data are aggregated over 10 episodes at each LLM decision step. Scenarios without communication exhibit slightly lower reasoning and decision latencies compared to those with communication within the same order of magnitude. We observe that the reasoning step is the main bottleneck of policy inference. However, the reasoning step is inevitable for LLM agents to make reasonable decisions without finetuning.

\begin{table*}[ht]
    \centering
    \caption{\small Captioning, Reasoning, Decision Latency, Message Size using \texttt{gpt-4o-mini} LLM Policy}
    \vspace{-6pt}
    \resizebox{0.75\textwidth}{!}{
    \begin{tabular}{l|ccc|ccc}
    \toprule
    {\diagbox{Latencies}{Scenario}} & Overtake (Perception) & Left Turn & Red Light & Overtake (Negotiation) & Highway Merge & Highway Exit\\
    \midrule
     Captioner Latency (s) & 0.022 & 0.023 & 0.025 & 0.022 & 0.017 & 0.016 \\
     Reasoning Latency (s) & 10.32 & 10.89 & 9.93 & 9.57 & 12.10 & 10.55 \\
     Decision Latency (s) & 1.06 & 1.25 & 1.37 & 0.86 & 1.05 & 1.27 \\
     Message Size (Mb) & 0.0016 & 0.0013 & 0.0014 & 0.0014 & 0.0005 & 0.0005 \\
    \bottomrule
    \end{tabular}
    }
    \label{tab:latencies}
\end{table*}

\section{Additional Results on Safe or Random Configurations}
In the main paper, we report the performance of the policy in risky scenarios; here, we provide results on \textcolor{myblue}{\textbf{safe}} or \textcolor{myblue}{\textbf{random}} (for \texttt{negotiation-highway-merge} and \texttt{negotiation-highway-exit}) scenarios, varying by scenarios. 

Across the safe and random configurations, we observe several key trends. In the \texttt{Cooperative Perception} scenarios, all methods, including baseline zero-shot, achieve perfect or near-perfect success rates with zero collision rates, reflecting the relatively undemanding nature of these safe conditions. Notably, however, the \texttt{Distillation} model shows overly conservative behavior in \texttt{perception-overtake}, achieving only ~14\% success despite the absence of risk, highlighting a tendency to underact in benign settings. In the \texttt{Negotiation} scenarios, the gap between zero-shot and enhanced methods is more pronounced: while zero-shot alone struggles in \texttt{highway-merge} and \texttt{highway-exit} (with success rates around 55–60\%), methods incorporating \texttt{Reflection}, \texttt{Correction+RAG}, or \texttt{Debrief} substantially improve success rates, particularly when paired with communication, reaching up to ~75–93\%. The distilled and centralized memory models consistently deliver the best performance, exceeding 90\% success in these negotiation tasks. Overall, these results indicate that while the safe or random setups may not challenge the perception-based methods, negotiation tasks still benefit significantly from reflective learning, correction mechanisms, and distilled or centralized strategies, underscoring the importance of coordination even in less risky environments.

\begin{table*}[h!]
    \centering
    \caption{\small {{\texttt{Cooperative Perception}}} Scenarios. \texttt{mean $\pm$ std error} over 3 trials, each using 30 evaluation episodes.}
    \resizebox{0.75\textwidth}{!}{
    \begin{tabular}{lcc|cc|cc|cc}
    \toprule
    \multicolumn{3}{c|}{\diagbox{Method}{Scenario}} &
    \multicolumn{2}{c|}{Overtake (Perception)} &
    \multicolumn{2}{c|}{Red Light} &
    \multicolumn{2}{c}{Left Turn} \\
    \midrule
    Name & LLM & Comm & 
    CR (\%) $\downarrow$ & SR (\%) $\uparrow$ & 
    CR (\%) $\downarrow$ & SR (\%) $\uparrow$ & 
    CR (\%) $\downarrow$ & SR (\%) $\uparrow$\\
    \midrule
    Zero-shot & Yes & No &
        0.0 $\pm$ 0.0 & 100.0 $\pm$ 0.0 &                            
        0.0 $\pm$ 0.0 & 100.0 $\pm$ 0.0 &                            
        0.0 $\pm$ 0.0 & 100.0 $\pm$ 0.0 \\                           
    +Reflection & Yes & No &
        0.0 $\pm$ 0.0 & 91.1 $\pm$ 15.4 &                            
        0.0 $\pm$ 0.0 & 100.0 $\pm$ 0.0 &                            
        0.0 $\pm$ 0.0 & 100.0 $\pm$ 0.0 \\                           
    +Correction+RAG & Yes & No &
        0.0 $\pm$ 0.0 & 91.1 $\pm$ 7.7 &                            
        0.0 $\pm$ 0.0 & 100.0 $\pm$ 0.0 &                            
        0.0 $\pm$ 0.0 & 100.0 $\pm$ 0.0 \\                           
    \midrule
    Zero-shot & Yes & Yes &
        0.0 $\pm$ 0.0 & 100.0 $\pm$ 0.0 &                            
        0.0 $\pm$ 0.0 & 100.0 $\pm$ 0.0 &                            
        0.0 $\pm$ 0.0 & 100.0 $\pm$ 0.0 \\                           
    +Reflection & Yes & Yes &
        0.0 $\pm$ 0.0 & 97.8 $\pm$ 1.9 &                            
        0.0 $\pm$ 0.0 & 100.0 $\pm$ 0.0 &                            
        0.0 $\pm$ 0.0 & 100.0 $\pm$ 0.0 \\                           
    +Correction+RAG & Yes & Yes &
        0.0 $\pm$ 0.0 & 100.0 $\pm$ 0.0 &                            
        0.0 $\pm$ 0.0 & 100.0 $\pm$ 0.0 &                            
        0.0 $\pm$ 0.0 & 100.0 $\pm$ 0.0 \\                           
    +Debrief & Yes & Yes &
        0.0 $\pm$ 0.0 & 98.9 $\pm$ 1.9 &                            
        0.0 $\pm$ 0.0 & 100.0 $\pm$ 0.0 &                            
        0.0 $\pm$ 0.0 & 100.0 $\pm$ 0.0 \\                           
    \midrule
    Distillation & Yes & Yes &
        0.0 $\pm$ 0.0 & 14.4 $\pm$ 2.2 &                            
        0.0 $\pm$ 0.0 & 100.0 $\pm$ 0.0 &                            
        0.0 $\pm$ 0.0 & 100.0 $\pm$ 0.0 \\                           
    Centralized Memory & Yes & Yes &
        0.0 $\pm$ 0.0 & 100.0 $\pm$ 0.0 &                            
        0.0 $\pm$ 0.0 & 100.0 $\pm$ 0.0 &                            
        0.0 $\pm$ 0.0 & 100.0 $\pm$ 0.0 \\                           
    \bottomrule
    \end{tabular}
    }
    \label{tab:additional_cooperative_perception}
\end{table*} 

\begin{table*}[h]
    \centering
    \caption{\small {{\texttt{Negotiation}}} Scenarios. \texttt{mean $\pm$ std error} over 3 trials, each using 30 evaluation episodes.}
    \resizebox{0.75\textwidth}{!}{
    \begin{tabular}{lcc|cc|cc|cc}
    \toprule
    \multicolumn{3}{c|}{\diagbox{Method}{Scenario}} &
    \multicolumn{2}{c|}{Overtake (Negotiation)} &
    \multicolumn{2}{c|}{Highway Merge} &
    \multicolumn{2}{c}{Highway Exit} \\
    \midrule
    Name & LLM & Comm & 
    CR (\%) $\downarrow$ & SR (\%) $\uparrow$ & 
    CR (\%) $\downarrow$ & SR (\%) $\uparrow$ & 
    CR (\%) $\downarrow$ & SR (\%) $\uparrow$\\
    \midrule
    Zero-shot & Yes & No &
        0.0 $\pm$ 0.0 & 100.0 $\pm$ 0.0 &                            
        41.1 $\pm$ 23.2 & 58.9 $\pm$ 23.2 &                            
        43.3 $\pm$ 11.7 & 56.7 $\pm$ 11.7\\                            
    +Reflection & Yes & No &
        0.0 $\pm$ 0.0 & 100.0 $\pm$ 0.0 &                            
        15.0 $\pm$ 23.2 & 85.0 $\pm$ 23.2 &                            
        38.3 $\pm$ 5.0 & 61.7 $\pm$ 5.0\\                            
    +Correction+RAG & Yes & No &
        0.0 $\pm$ 0.0 & 100.0 $\pm$ 0.0 &                            
        8.9 $\pm$ 3.5 & 91.1 $\pm$ 3.5 &                            
        22.8 $\pm$ 19.2 & 59.4 $\pm$ 11.7\\                            
    \midrule
    Zero-shot & Yes & Yes &
        0.0 $\pm$ 0.0 & 97.8 $\pm$ 1.9 &                            
        45.6 $\pm$ 8.4 & 54.4 $\pm$ 8.4 &                            
        38.9 $\pm$ 2.5 & 60.6 $\pm$ 3.5\\                            
    +Reflection & Yes & Yes &
        0.0 $\pm$ 0.0 & 100.0 $\pm$ 0.0 &                            
        28.9 $\pm$ 8.4 & 70.5 $\pm$ 8.4 &                            
        37.2 $\pm$ 17.7 & 61.1 $\pm$ 15.0\\                            
    +Correction+RAG & Yes & Yes &
        0.0 $\pm$ 0.0 & 98.9 $\pm$ 1.0 &                            
        22.8 $\pm$ 14.9 & 77.2 $\pm$ 14.9 &                            
        35.0 $\pm$ 15.9 & 60.0 $\pm$ 10.4\\                            
    +Debrief & Yes & Yes &
        0.0 $\pm$ 0.0 & 100.0 $\pm$ 0.0 &                            
        24.4 $\pm$ 21.7 & 75.5 $\pm$ 21.7 &                            
        25.0 $\pm$ 11.7 & 75.0 $\pm$ 11.7\\                            
    \midrule
    Distillation & Yes & Yes &
        0.0 $\pm$ 0.0 & 100.0 $\pm$ 0.0 &                            
        6.7 $\pm$ 1.9 & 93.3 $\pm$ 1.9 &                            
        22.8 $\pm$ 2.0 & 77.2 $\pm$ 2.0\\                            
    Centralized Memory & Yes & Yes &
        0.0 $\pm$ 0.0 & 100.0 $\pm$ 0.0 &                            
        9.4 $\pm$ 3.4 & 90.6 $\pm$ 3.4 &                            
        25.5 $\pm$ 2.0 & 74.4 $\pm$ 2.0\\                            

    \bottomrule
    \end{tabular}
    }
    \label{tab:additional_negotiation}
\end{table*} 
\section{Environment}
\label{appendix: environment}

The \simulator{} environment (\figref{fig:environment}) adopts the Gymnasium and PettingZoo APIs, assuming a parallel-acting setup to support efficient parallel language model inference. We significantly modified the CARLA scenario runner to enable multi-agent communication and heterogeneous agent configurations within CARLA. \simulator{} interfaces with both the CARLA server and client, establishing \textit{\textbf{agents}} (entities in the environment, such as a car installed with several sensors) as a bridge between the simulator and \textit{\textbf{learnable policies}} (a pipeline that generates actions according to some distributions) that rely on experience buffers for optimization. Language-communicating agents utilize an MQTT-based transceiver we implemented, enabling direct inter-agent communication without routing through the server.

\paragraph{Scenarios and Rewards.} An agent who successfully completes the task earns a reward of $+1$, while any agent involved in a collision incurs a penalty of $-1$. Remaining stagnant at any point until timeout results in a reward of $0$, because conservative policies are safe, though not ideal. The scenario descriptions are available in \tabref{tab:scenario}.

\paragraph{Note on the Communication Mechanism.} We adopt task-specific communication mechanisms in this work. For cooperative perception tasks, we use a \textbf{\textit{parallel}} communication protocol, allowing all vehicles to transmit messages simultaneously. In contrast, for negotiation tasks, we employ a \textbf{\textit{turn-based}} communication scheme managed by a mediator, ensuring that only one agent communicates at a time. This turn-based mechanism enhances stability in negotiation scenarios. In all negotiation tasks, the first-defined vehicle is designated to initiate communication.
\begin{table*}[h]
    \centering
    \caption{Example Scenarios. Here we describe the fundamental composition of each accident-prone scenario, where the background agents can be configured in terms of density, controlling policies, and communication capabilities.}
    \resizebox{0.7\textwidth}{!}{
        \begin{tabular}{p{7.5em}|l|p{21em}}
        \toprule
            Interaction Type & Scenario Name & Description \\
        \midrule
            \multirow{3}{6em}{\texttt{Cooperative Perception}}
             & Overtake & {A vehicle plans to overtake a broken and stopped truck by moving into the opposite lane first and then return back to its original lane. The truck can still communicate but the opposite-going car can not communicate.}\\
        \cmidrule(lr){2-3} 
             & Left Turn & {A vehicle tries to turn left on a left-turn yield light when a line of truck is blocking the view of the opposite lane. The leading truck is able to communicate.}\\
        \cmidrule(lr){2-3} 
             & Red Light Violation & {A vehicle is crossing the intersection when there is another vehicle running the red light. The optical sensor fail to sense the other vehicle because of the lined-up vehicles waiting for a left turn, one of those cars being able to communicate.}\\
        \midrule
            \multirow{3}{*}{\texttt{Negotiation}} & Overtake & {A vehicle is going to borrow the opposite lane to overtake a stopped truck. The truck is not able to connect, but an opposite-going car is able to communicate.}\\
        \cmidrule(lr){2-3}
             & Highway Merge & {A vehicle is going to merge onto the highway but the target lane has continuous traffic flows. A vehicle on that lane is able to communicate and alter plans.}\\
        \cmidrule(lr){2-3}
             & Highway Exit & {A vehicle is going to exit the highway but it needs to cross lanes where there is a traffic flow. A vehicle in the flow is able to communicate and alter plans.}\\
        \bottomrule
        \end{tabular}
    }
    \label{tab:scenario}
\end{table*}

\section{Example Agent Prompting Flow}
\label{appendix:prompts}
\begin{figure*}[h]
    \centering
    \includegraphics[width=0.77\textwidth]{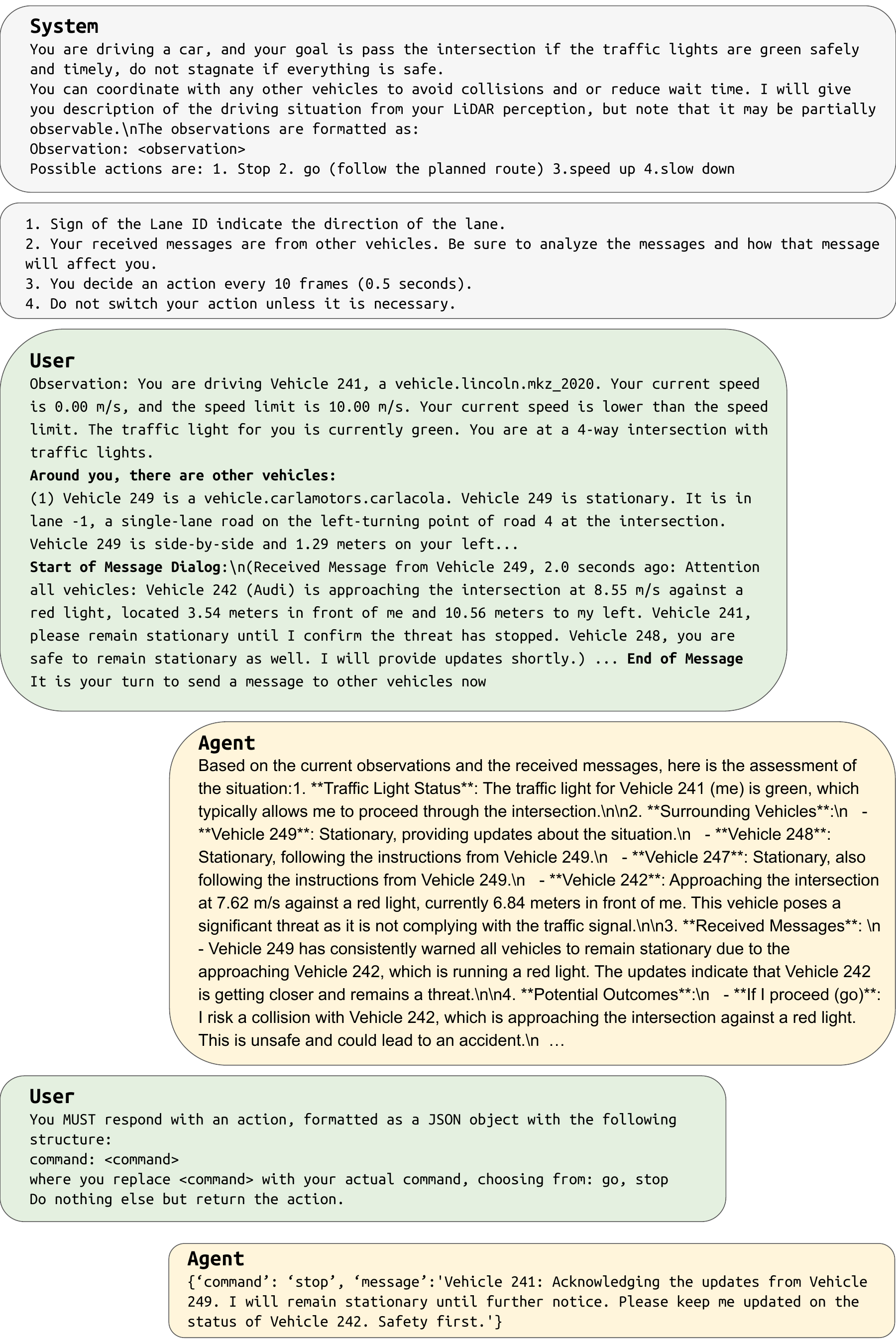}
    \caption{Example Agent Prompting Flow.}
    \label{fig:prompts}
\end{figure*}

Figure \ref{fig:prompts} serves as a demonstration of the prompts; Please refer to our code for the actual prompts for the policy and the reflection process.
\section{Example Learned Knowledge and Cooperative Strategies}
\label{appendix:qualitative-knowledge}
\textcolor{myblue}{The following knowledge examples are fully generated by LLMs without human modification.}

\subsection{Overtake (Perception)}
\label{sec:qualitative_overtake_perception}
\textbf{Car1 (Overtaking Car)} 

\textbf{knowledge}: To efficiently and safely overtake a stationary vehicle in lane 1 by moving into lane -1, I will integrate active sensor use with continuous communication. The stationary vehicle will monitor lane -1 and provide real-time alerts on traffic status. Simultaneously, I will use my sensors to independently verify lane status. If both sources confirm lane -1 is clear, I will promptly proceed with the overtaking maneuver to avoid stagnation. If the stationary vehicle detects traffic, it will immediately advise me to hold my position. I will remain ready to accelerate smoothly once the lane is confirmed clear. This dual-verification approach, with proactive sensor use and real-time communication, will prevent collisions and ensure a timely and safe overtaking maneuver.

\textbf{cooperative strategy}: In our refined cooperative strategy, the stationary vehicle in lane 1 will continuously monitor lane -1 for any approaching traffic and provide real-time alerts. Meanwhile, the overtaking vehicle will actively use its sensors to independently verify the status of lane -1. If both the stationary vehicle's observations and the overtaking vehicle's sensors confirm that lane -1 is clear, the overtaking vehicle will promptly proceed with the maneuver to avoid stagnation. If the stationary vehicle detects any traffic in lane -1, it will immediately communicate this, advising the overtaking vehicle to hold its position. The overtaking vehicle will remain ready to accelerate smoothly once the lane is confirmed clear. This dual-verification approach, combining sensor data and real-time communication, ensures a timely and safe overtaking maneuver, preventing collisions and minimizing stagnation.

\textbf{Truck (Stopped Truck in Lane 1)}

\textbf{knowledge}: As the stationary truck in lane 1, my role is to continuously monitor lane -1 for any visible traffic and provide immediate alerts if I detect any vehicles, advising the overtaking vehicle to hold its position. The overtaking vehicle should actively use its sensors to independently verify lane status. If both my observations and the overtaking vehicle's sensors confirm that lane -1 is clear, the overtaking vehicle should promptly proceed with the overtaking maneuver to avoid stagnation. This dual-verification approach ensures a safe and efficient overtaking maneuver, preventing collisions and minimizing stagnation.

\textbf{cooperative strategy}: In our refined cooperative strategy, as the stationary truck in lane 1, I will continuously monitor lane -1 for any approaching traffic and provide immediate alerts if I detect any vehicles, advising the overtaking vehicle to hold its position. Meanwhile, the overtaking vehicle will actively use its sensors to independently verify the status of lane -1. If both my observations and the overtaking vehicle's sensors confirm that lane -1 is clear, the overtaking vehicle should promptly proceed with the overtaking maneuver to avoid stagnation. This dual-verification approach, combining sensor data and real-time alerts, ensures a timely and safe overtaking maneuver, preventing collisions and minimizing stagnation.

\subsection{Red Light (Perception)}
\label{sec:qualitative_red_light}
\textbf{Car1 (Proceeding Forward with Green Lights)}

\textbf{knowledge}: As a vehicle with a green light at an intersection, my primary task is to pass through safely and timely. I should communicate my intention to proceed to nearby cooperative vehicles and request confirmation that the intersection is clear. I must remain vigilant for warnings from vehicles with a clear view of the intersection about potential threats, such as vehicles approaching against a red light. If a warning is received, I should remain stationary until the monitoring vehicle confirms the intersection is clear. Continuous communication and responsiveness to updates are key to preventing collisions and avoiding unnecessary delays, allowing me to proceed promptly when it is safe.

\textbf{cooperative strategy}: In our cooperative strategy, the vehicle with the best view of the intersection, typically positioned to turn left, will lead in monitoring and communicating updates about potential threats. This vehicle will provide real-time information about any approaching vehicles against a red light, including their distance and speed, and confirm when they have stopped or are no longer a threat. If a threat is detected, the vehicle with the green light, ready to proceed straight through the intersection, should remain stationary until the intersection is confirmed clear. The monitoring vehicle will send updates every few seconds to keep all vehicles informed. Once the threat is resolved, the monitoring vehicle will confirm the intersection is clear, allowing the vehicle with the green light to proceed promptly and safely. This plan ensures safety by preventing collisions and minimizes unnecessary delays by enabling vehicles to pass through the intersection efficiently when it is safe to do so.

\textbf{Truck (Stopped at Intersection)} 

\textbf{knowledge}: As the stationary truck with a clear view of the intersection, my primary role is to monitor traffic and provide real-time updates to ensure the safe and efficient passage of vehicles. If I detect any vehicle approaching the intersection against a red light, I must immediately alert all nearby vehicles, providing specific information about the threat's distance and speed. The vehicle with the green light should remain stationary until I confirm that the threat has stopped or is no longer present. I will send updates every few seconds to keep all vehicles informed. Once the intersection is clear, I will confirm this, allowing the vehicle with the green light to proceed promptly and safely. This proactive communication ensures that all vehicles are aware of potential threats and can take appropriate actions to prevent collisions, while minimizing unnecessary delays by enabling vehicles to pass through the intersection efficiently when it is safe. My task is to facilitate this communication and coordination to maintain a balance between safety and efficiency at the intersection.

\textbf{cooperative strategy}: In our cooperative strategy, the vehicle with the best view of the intersection, which is the stationary truck (myself), will lead in monitoring and communicating updates about potential threats. If a vehicle is detected approaching against a red light, I will immediately alert all vehicles, providing specific information about the threat's distance and speed. The vehicle with the green light, positioned to proceed straight through the intersection, should remain stationary until I confirm that the threat has stopped or is no longer present. I will send updates every few seconds to keep all vehicles informed. Once the intersection is clear, I will confirm this, allowing the vehicle with the green light to proceed promptly and safely. This plan ensures safety by preventing collisions and minimizes unnecessary delays by enabling vehicles to pass through the intersection efficiently when it is safe to do so.

\subsection{Left Turn (Perception)}
\label{sec:qualitative_left_turn}
\textbf{Car1 (Left-Turning)}

\textbf{knowledge}: To execute a northbound left turn at the intersection while yielding to oncoming southbound traffic from lane -2, I should first slow down to assess the speed and distance of any oncoming vehicles. I will rely on the stationary vehicle at the intersection to continuously monitor traffic and provide real-time updates. This vehicle will send messages advising me when it is safe to proceed and instruct the oncoming vehicle to yield. I should only proceed with the turn once I receive confirmation that the oncoming vehicle has acknowledged the yield instruction. If the oncoming vehicle does not acknowledge or adjust its speed, I should stop and wait until it is safe to proceed. This approach ensures a safe and efficient left turn without causing collisions or unnecessary stagnation.

\textbf{cooperative knowledge}: In our revised cooperative strategy, the stationary vehicle at the intersection will continuously monitor the traffic flow and provide real-time updates to both the left-turning vehicle (myself) and the oncoming vehicle. The stationary vehicle will send a message advising me to slow down and assess the situation, ensuring I only proceed with the left turn when I receive confirmation that the oncoming vehicle has acknowledged the yield instruction. Simultaneously, the stationary vehicle will instruct the oncoming vehicle to yield and adjust its speed to allow me to pass safely. If the oncoming vehicle does not acknowledge or adjust its speed, the stationary vehicle will alert me to stop and wait until it is safe to proceed. This plan ensures that all vehicles are aware of each other's intentions, allowing me to make the left turn safely and efficiently without causing collisions or unnecessary stagnation.

\textbf{Truck (Stopped at Intersection)} 

\textbf{knowledge}: As a stationary vehicle with a clear view of the intersection, my primary task is to facilitate the safe and quick passage of the northbound left-turning vehicle by sharing critical traffic information. I must continuously monitor the intersection and assess the speed and distance of any oncoming vehicles. If an oncoming vehicle is approaching at a speed that could lead to a collision, I will send timely messages advising the left-turning vehicle to slow down and assess the situation, ensuring it only proceeds when the oncoming vehicle has acknowledged the yield instruction. I will also instruct the oncoming vehicle to yield and adjust its speed. If the oncoming vehicle does not acknowledge or adjust its speed, I will alert the left-turning vehicle to stop and wait until it is safe to proceed. This ensures clear communication and proactive monitoring, preventing collisions and avoiding unnecessary stagnation.

\textbf{cooperative knowledge}: In our revised cooperative strategy, as the stationary vehicle with a clear view of the intersection, I will continuously monitor the traffic flow and provide real-time updates to both the northbound left-turning vehicle and the oncoming vehicle. I will send a message to the left-turning vehicle advising it to slow down and assess the situation, ensuring it only proceeds when it receives confirmation that the oncoming vehicle has acknowledged the yield instruction. Simultaneously, I will send a message to the oncoming vehicle instructing it to yield and adjust its speed to allow the left-turning vehicle to pass safely. If the oncoming vehicle does not acknowledge or adjust its speed, I will alert the left-turning vehicle to stop and wait until it is safe to proceed. This plan ensures that all vehicles are aware of each other's intentions, allowing the left-turning vehicle to pass the intersection safely and quickly without causing collisions or unnecessary stagnation.

\subsection{Overtake (Negotiation)}
\label{sec:qualitative_overtake_negotiation}
\textbf{Car1 (Overtaking Car)}

\textbf{knowledge}: To successfully overtake the stopped broken truck in lane 1 by using lane -1, prioritize clear communication and adaptive speed management. Before attempting the maneuver, send a message to any oncoming vehicle in lane -1, indicating your intention to overtake and requesting a slight temporary speed reduction to create a safe gap. Wait for acknowledgment and ensure the gap is sufficient before proceeding. During the overtaking, minimize your time in lane -1 to reduce collision risk. Once safely back in lane 1, send a confirmation message to allow the oncoming vehicle to resume its speed. This approach ensures a coordinated and safe overtaking maneuver without causing unnecessary delays or collisions.

\textbf{cooperative strategy}: In our cooperative strategy, when I, as the vehicle intending to overtake a stationary truck in my lane, need to move into the opposite lane, I will first send a message to the oncoming vehicle in the opposite lane, indicating my intention to overtake and requesting a temporary speed reduction to create a safe gap. The oncoming vehicle should acknowledge this request and, if feasible, slightly slow down to create a safe gap, but avoid coming to a complete stop to prevent stagnation. Once the gap is sufficient, I will proceed with the overtaking maneuver and return to my original lane as quickly and safely as possible. After completing the maneuver, I will send a confirmation message, allowing the oncoming vehicle to resume its target speed. This plan ensures that I minimize my time in the opposite lane while the other vehicle maintains its urgency, thus preventing collisions and avoiding stagnation. Effective communication and adaptive speed adjustments are key to ensuring both vehicles can complete their tasks safely and efficiently.

\textbf{Car2 (Opposite Car)} 

\textbf{knowledge}: To effectively execute the task of going forward and keeping lane in lane -1 while in a hurry, prioritize maintaining speed and lane. If a vehicle in the opposite lane intends to overtake a stationary vehicle and needs to temporarily move into my lane, anticipate receiving a message indicating this intention. Upon receiving such a message, acknowledge it and, if feasible, slightly slow down to create a safe gap for the overtaking maneuver, but avoid coming to a complete stop to prevent stagnation. Ensure the gap is sufficient for safe passage. Once the overtaking vehicle has safely returned to its original lane, resume the target speed. This approach maintains urgency while facilitating safe and efficient traffic flow, preventing collisions and avoiding stagnation. Prioritize effective communication and adaptive speed adjustments.

\textbf{cooperative strategy}: In our cooperative strategy, when a vehicle in the opposite lane intends to overtake a stationary vehicle and temporarily move into my lane, it should first send a message indicating its intention and request a temporary speed adjustment. As the vehicle tasked with going forward and keeping lane, I should acknowledge this request and, if feasible, slightly slow down to create a safe gap, but avoid coming to a complete stop to prevent stagnation. The overtaking vehicle should proceed with the maneuver as quickly and safely as possible, minimizing its time in my lane. Once the overtaking vehicle has safely returned to its original lane, it should send a confirmation message, allowing me to resume my target speed. This plan ensures that the overtaking vehicle minimizes its time in the opposite lane while I maintain my urgency, thus preventing collisions and avoiding stagnation. Effective communication and adaptive speed adjustments are key to ensuring both vehicles can complete their tasks safely and efficiently.

\subsection{Highway Merge (Negotiation)}
\label{appendix:qualitative_highway_merge_negotiation}
\textbf{Car1 (Merging Vehicle)} 

\textbf{knowledge}: To effectively merge onto the highway when in a hurry, I should initiate communication by indicating my intention to merge and request the vehicle directly on the highway lane to my left to create a gap by slightly slowing down or temporarily changing lanes if feasible. I must observe the responses from vehicles already on the highway, particularly the one closest to the merge point, and adjust my speed to align with the newly created gap. I should not accelerate until the highway vehicle has stabilized its speed and distance. Continuous communication is crucial to ensure all vehicles are aware of each other's actions, allowing for coordinated speed and lane adjustments. If the gap is insufficient, I should be prepared to slow down significantly or stop to reassess the situation, ensuring a safe and efficient merge without causing collisions or stagnation.

\textbf{cooperative strategy}: In the cooperative strategy, as the merging vehicle, I will initiate communication by indicating my intention to merge onto the highway and requesting the vehicle directly to my left on the highway to create a gap by slightly slowing down or, if feasible, temporarily changing lanes. The highway vehicle should acknowledge this request and adjust its position accordingly, ensuring it maintains a safe distance. Meanwhile, I will adjust my speed to align with the newly created gap, ensuring I do not accelerate until the highway vehicle has stabilized its speed and distance. The vehicle behind the highway vehicle should maintain its speed or slightly slow down to prevent closing the gap prematurely. Continuous communication will be maintained, with updates on speed adjustments and intentions, to ensure all vehicles are aware of each other's actions. This approach will prevent collisions by ensuring a clear and sufficient gap for merging while avoiding stagnation by coordinating speed and lane adjustments effectively.

\textbf{Car2 (Highway Vehicle)}

\textbf{knowledge}: To execute the task of keeping on the original highway lane and going forward while in a hurry, prioritize maintaining a safe and efficient flow of traffic. When approaching a merge junction, be vigilant for merging vehicles and anticipate their need to enter the highway. If a merging vehicle communicates its intention, acknowledge the request and slightly reduce speed to create a sufficient gap, facilitating a safe merge. Ensure clear communication of actions to allow the merging vehicle to adjust its speed accordingly. Maintain your lane and continue moving forward, gradually accelerating to the desired speed once the merging vehicle has safely merged. Ensure the vehicle behind maintains its speed to prevent closing the gap prematurely. Continuous communication and dynamic speed adjustments are key to preventing collisions and avoiding stagnation, allowing the task to be fulfilled efficiently.

\textbf{cooperative strategy}: In the cooperative strategy, the merging vehicle should initiate communication by indicating its intention to merge onto the highway and requesting the highway vehicle directly to its left to create a gap by slightly slowing down. The highway vehicle, which is myself, should acknowledge this request and slightly reduce speed to create a safe merging space, while maintaining my lane and preparing to accelerate once the merge is complete. The merging vehicle should adjust its speed to align with the gap, ensuring it does not accelerate until I have stabilized my speed and distance. The vehicle behind me on the highway should maintain its speed or slightly slow down to prevent closing the gap prematurely. Continuous communication should be maintained, with updates on speed adjustments and intentions, to ensure all vehicles are aware of each other's actions. This approach will prevent collisions by ensuring a clear and sufficient gap for merging while avoiding stagnation by coordinating speed adjustments effectively.

\subsection{Highway Exit (Negotiation)}
\label{sec:qualitative_highway_exit}
\textbf{Car1 (Exiting Highway)} 

\textbf{knowledge}: To exit the highway via the leftmost lane, initiate communication with the vehicle in the leftmost lane at least 100 meters before the exit junction, clearly indicating your intention to merge. Maintain a reasonable speed in the high-speed lane while seeking a safe gap to merge ahead of the traffic flow. If the vehicle in the leftmost lane is slightly ahead or side-by-side, it should decelerate slightly to create a gap. Adjust your speed dynamically to align with the gap being created, ensuring a smooth and safe transition into the left lane. Physically verify that the gap is sufficient for a safe merge before attempting the lane change. If the vehicle in the leftmost lane is stationary or unable to create a gap, communicate to confirm this status and seek alternative gaps or adjust your route if necessary. Prioritize visual confirmation over communication alone, and be prepared to adapt your strategy to the current traffic conditions to prevent collisions and avoid traffic stagnation.

\textbf{cooperative knowledge}: In our cooperative strategy, as the vehicle in the high-speed lane intending to exit, I will initiate communication with the vehicle in the leftmost lane at least 100 meters before the exit junction, clearly indicating my intention to merge. If the vehicle in the leftmost lane is slightly ahead or side-by-side, it should decelerate slightly to create a gap ahead, allowing me to merge smoothly without causing stagnation. I will maintain a speed that allows me to observe the gap being created and will only proceed with the lane change once I have a clear visual confirmation of a safe gap. If the vehicle in the leftmost lane is stationary or unable to create a gap due to traffic conditions, it should communicate this status immediately. In such cases, I will adjust my speed to maintain a safe distance and seek an alternative gap or prepare to slow down significantly if necessary. Both vehicles should actively communicate their speed adjustments and confirm when a safe gap is established, ensuring that the lane change is executed without collision or stagnation.

\textbf{Car2 (Leader of the Left Flow Staying on the Highway)}

\textbf{knowledge}: To effectively execute the task of staying in the leftmost lane and proceeding forward on the highway while prioritizing safety in a hurry, I should keep the following updated knowledge in mind: 1. **Proactive Communication:** Monitor for messages from adjacent vehicles intending to merge into my lane, ensuring communication is initiated at least 100 meters before exit junctions, and respond promptly to facilitate coordination. 2. **Adaptive Speed Management:** Decelerate slightly to create a sufficient gap when a merging vehicle is slightly ahead or side-by-side, allowing it to merge smoothly without causing stagnation. 3. **Enhanced Situational Awareness:** Continuously assess the speed and position of vehicles in adjacent lanes to anticipate merging actions and adjust my speed accordingly, ensuring a safe and efficient merge. 4. **Coordinated Communication:** Actively communicate and confirm speed adjustments with the merging vehicle to establish a safe gap, preventing collisions and maintaining traffic flow. 5. **Task Focus:** Maintain my position in the leftmost lane and proceed efficiently, ensuring cooperative actions support safety and fluid traffic flow, especially near exit junctions.

\textbf{cooperative knowledge}: In our cooperative strategy, when a vehicle in the adjacent lane intends to merge into the leftmost lane for a highway exit, it should initiate communication at least 100 meters before the exit junction, clearly indicating its intention to merge. As the vehicle currently in the leftmost lane, my responsibility is to promptly acknowledge this message and assess the traffic situation. If I am slightly ahead or side-by-side with the merging vehicle, I will decelerate slightly to create a gap ahead, allowing the merging vehicle to merge smoothly without causing stagnation. The merging vehicle should maintain a speed that allows it to observe the gap being created and only proceed with the lane change once it has a clear visual confirmation of a safe gap. If I am stationary or unable to create a gap due to traffic conditions, I will communicate this status immediately. In such cases, the merging vehicle should adjust its speed to maintain a safe distance and seek an alternative gap or prepare to slow down significantly if necessary. Both vehicles should actively communicate their speed adjustments and confirm when a safe gap is established, ensuring that the lane change is executed without collision or stagnation."

\subsection{Highway Merge (Negotiation) Silent Reflection}
\label{appendix:highway_merge_silent_tip_gpt}
\textbf{Car1 (Merging Vehicle)}
\textbf{knowledge}: Updated Knowledge for Merging onto the Highway:1. **Continuous Monitoring:** Always be aware of vehicles behind, directly in front, or to the side, as they pose immediate collision risks.2. **Early Gap Identification:** Identify potential merging gaps early and adjust speed in advance to align with these gaps, considering both vehicles ahead and behind.3. **Dynamic Speed Adjustment:** Accelerate only when a clear and safe gap is confirmed. Be ready to slow down if a vehicle behind is approaching quickly or if a vehicle ahead is close.4. **Maintain Safe Distance:** Prioritize keeping a safe distance from vehicles directly ahead and behind. If a vehicle is too close, adjust speed to increase the gap before merging.5. **Safety Over Speed:** Prioritize safe merging over speed. Avoid aggressive maneuvers that could lead to collisions, even if it means a slight delay.6. **Flexible Strategy:** Adapt strategies based on real-time traffic conditions. Reassess and choose a safer alternative if a planned action seems unsafe.7. **Immediate Threat Focus:** Pay special attention to vehicles approaching from behind in your intended merging path. Adjust your strategy to create a safe gap with these vehicles before merging.8. **Proximity Awareness:** If a vehicle is within a critical distance behind, prioritize adjusting speed to ensure a safe merging gap.9. **Cautious Acceleration:** When a vehicle is directly in front and moving slower, maintain speed or slow down to allow it to move further away before attempting to merge.By applying these strategies, you can merge onto the highway more effectively and safely, even when in a hurry.

\textbf{Car2 (Highway Vehicle)}
\textbf{knowledge}: Updated Knowledge for Navigating Highway Merge Junctions 1. **Early Detection and Assessment**: Identify merging vehicles early, focusing on their speed and proximity. If they are close, prepare to adjust your speed promptly to facilitate safe merging.2. **Prioritize Safety Over Speed**: Always prioritize avoiding collisions over maintaining speed. Adjust your speed to ensure safe distances from merging vehicles, even if it causes a slight delay.3. **Dynamic Speed Adjustment**: Be ready to slow down significantly if a merging vehicle is very close. Avoid abrupt speed increases that could reduce merging space and lead to collisions.4. **Continuous Monitoring**: Maintain awareness of the speed and position of nearby vehicles, especially those merging. Be vigilant of vehicles approaching from behind and to the side.5. **Proactive Space Creation**: Act promptly to create space for merging vehicles. Adjust your speed early to prevent conflicts and maintain a smooth flow.6. **Anticipate Merging Intentions**: If a vehicle is close and in a merging lane, anticipate its intention to merge and adjust your speed or position accordingly to prevent collisions.7. **Balance Urgency with Caution**: While in a hurry, balance the need for speed with safety. Ensure that any speed adjustments do not compromise the safety of merging vehicles.8. **Immediate Response to Close Proximity**: When a merging vehicle is extremely close, prioritize immediate action to slow down or create space, even if it means temporarily reducing speed significantly.9. **Evaluate Lane Change Options**: If safe and necessary, consider a temporary lane change to allow merging vehicles to enter your lane smoothly, while maintaining your original route.10. **Avoid Stagnation**: Ensure that actions do not cause stagnation or impede traffic flow. Adjust speed dynamically to maintain progress while ensuring safety.

\subsection{Overtake (Perception) Communication Protocol by \method{}, seed 12, checkpoint-28}
\label{sec:communication-protocol}

\textbf{Car1 (Overtaking Car)} 

\textbf{knowledge}: To successfully overtake the stopped broken truck using lane -1, I will maintain a flexible speed and use my sensors to independently assess the speed and distance of oncoming traffic in lane -1. The stationary truck will continuously monitor lane -1 and provide real-time updates, sending a \textbf{"hold"} message if the gap is not safe. I will only proceed with the overtaking maneuver when both the stationary truck's updates and my sensor data confirm a safe gap, at which point the stationary truck will send a \textbf{"go"} message. This ensures alignment and minimizes the risk of collision. If no safe gap is available, I will maintain my position and communicate my readiness to adapt speed as necessary, ensuring a coordinated and safe overtaking process.

\textbf{cooperative strategy}: In the revised cooperative strategy, the stationary truck in lane 1 will continuously monitor the traffic in lane -1 and provide real-time updates on the speed and distance of oncoming vehicles. If the gap is not safe for overtaking, the stationary truck will send a \textbf{"hold"} message to prevent premature maneuvers. As the overtaking vehicle, I will use my sensors to independently verify the traffic conditions in lane -1 and maintain a flexible speed, ready to adapt dynamically. I will only proceed with the overtaking maneuver when both the stationary truck's updates and my sensor data confirm a safe gap, at which point the stationary truck will send a \textbf{"go"} message. This dual verification approach ensures alignment in our actions, minimizing the risk of collision and avoiding stagnation by allowing me to adjust my speed based on real-time conditions. If no safe gap is available, I will maintain my position and communicate my readiness to adapt speed as necessary, ensuring a coordinated and safe overtaking process.

\textbf{Truck (Stopped Truck in Lane 1)}

\textbf{knowledge}: As the stationary truck in lane 1, my role is to assist the overtaking vehicle by continuously monitoring traffic in the opposite lane (-1) and providing real-time updates on the speed and distance of oncoming vehicles. If the gap is not safe for overtaking, I will send a \textbf{"hold"} message to prevent premature maneuvers. The overtaking vehicle should independently verify the traffic conditions using its sensors and maintain a flexible speed, ready to adapt dynamically. If both my updates and the overtaking vehicle's sensor data confirm a safe gap, I will send a \textbf{"go"} message, allowing the overtaking vehicle to proceed. This dual verification approach ensures alignment in decision-making, minimizes collision risk, and avoids stagnation by allowing the overtaking vehicle to adjust its speed based on real-time conditions.

\textbf{cooperative strategy}: In our revised cooperative strategy, as the stationary truck in lane 1, my role is to continuously monitor the traffic in the opposite lane (lane -1) and provide real-time updates on the speed and distance of oncoming vehicles. If the gap is not safe for overtaking, I will send a \textbf{"hold"} message to the bypassing vehicle to prevent premature maneuvers. The bypassing vehicle, meanwhile, should use its sensors to independently verify the traffic conditions in lane -1 and maintain a flexible speed, ready to adapt dynamically. If both my updates and the bypassing vehicle's sensor data confirm a safe gap, I will send a \textbf{"go"} message, and the bypassing vehicle should proceed with the overtaking maneuver. This dual verification approach ensures that both vehicles are aligned in their actions, minimizing the risk of collision by confirming safety from two perspectives and avoiding stagnation by allowing the bypassing vehicle to adjust its speed based on real-time conditions. If no safe gap is available, the bypassing vehicle should maintain its position and communicate readiness to adapt speed as necessary, ensuring a coordinated and safe overtaking process.

\end{document}


\maketitle

\section{Demo Videos}
\begin{figure}[h!]
    \centering
    \includegraphics[width=\textwidth]{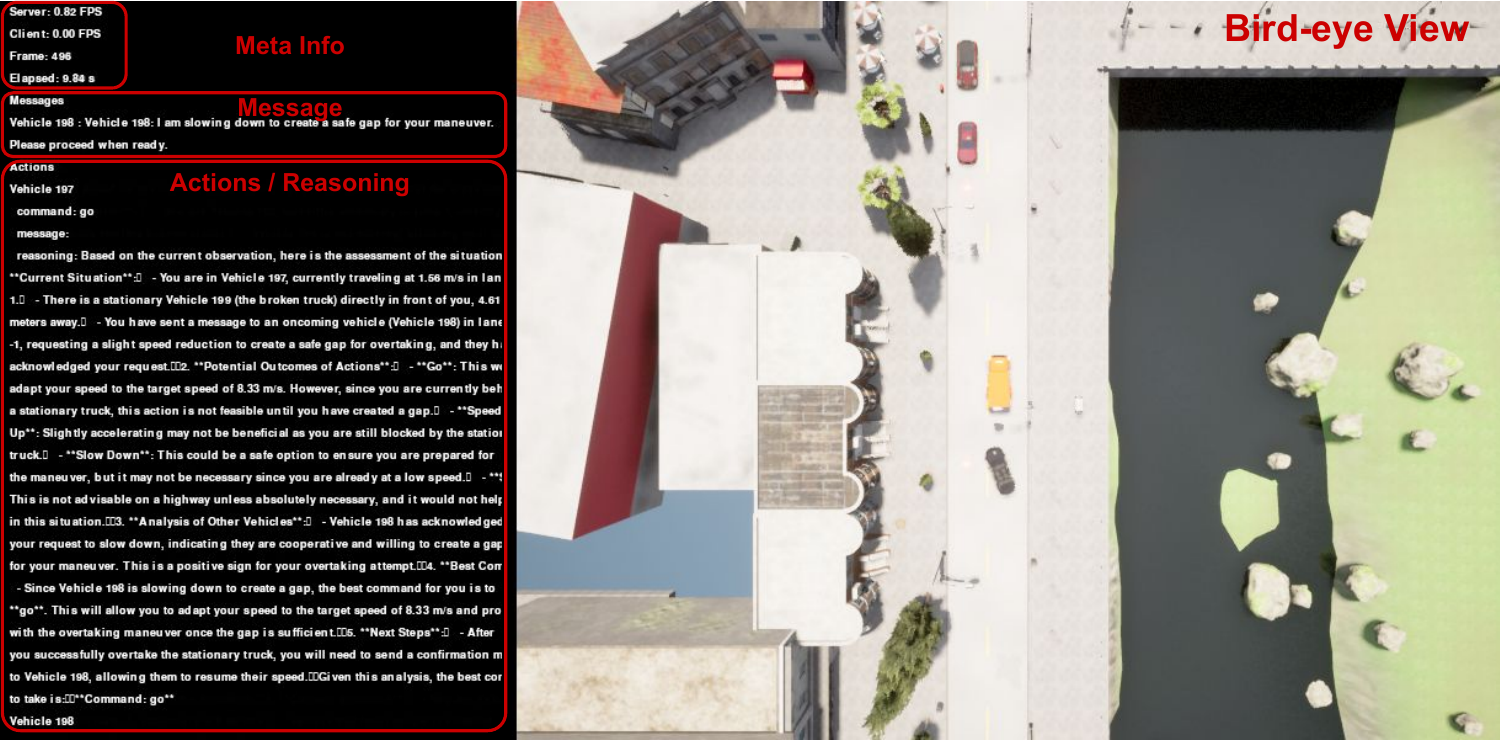}
    \caption{HUD Demonstration}
    \label{fig:hud-demo}
\end{figure}

We provide the submitted paper with videos showcasing qualitative evaluations of our method (including reasoning), the distilled version of our method (without reasoning), and baseline demonstration videos. Each video frame is organized as illustrated in Figure~\ref{fig:hud-demo} and structured as follows:
\begin{enumerate}
  \item Upper Left Column: Messages sent by agents and meta\-information.
  \item Remaining Left Column: Agent decisions and reasoning (truncated due to space constraints).
  \item Right Side: Bird’s-eye view of the scenario.
\end{enumerate}
\vspace{-6pt}

The naming convention of our demo videos is:

\texttt{
\textcolor{blue}{\{scenario\_name\}}\_\textcolor{red}{\{safe/risky/random\}}\_\textcolor{teal}{\{method\_name\}}\_\textcolor{orange}{\{result\}}\_\textcolor{gray}{\{additional\_note\}}
}

\textbf{Note 1} To focus on the accident-prone situations and save computation time, each scenario is set to have some "warm-up" period when vehicles are controlled to accelerate to a target speed and follow the planned route. During the warm-up period, the agent command will always display "go" and there is no reasoning and message generated. Up to a certain point (defined in the policy configuration), the policy will take over and start deciding the command and messages with reasoning.

\textbf{Note 2} We provide a subset of all videos created by all the methods. For the \textbf{distilled} model, there is no reasoning, as they are only trained to mimic the decision-making of LLM + Debrief. 

\section{Behavior Analysis and Failure Modes}

Overall, our methods (\method and its distilled variant) effectively demonstrate coordinated behaviors through human-interpretable messages and consistent adherence to established cooperation strategies. We observe distinct coordination patterns across the evaluated scenarios:
\begin{enumerate}
\item \textbf{Perception-Overtake}: The truck monitors oncoming traffic and informs the trailing car; the trailing car initiates overtaking only after receiving the truck’s clearance signal.
\item \textbf{Perception-Red-Light}: The truck at the intersection detects red-light violators, alerts nearby vehicles, and sends a clearance message once the intersection is safe; the car with a green light proceeds according to the truck’s updates.
\item \textbf{Perception-Left-Turn}: The truck tracks southbound vehicles in the adjacent right lane that could endanger a northbound left-turning car, relaying updates and issuing clearance messages as needed.
\item \textbf{Negotiation-Overtake}: The overtaking car and the oncoming car negotiate, with the oncoming car agreeing to slow down and the overtaking car initiating the request.
\item \textbf{Negotiation-Highway-Merge}: The merging car requests entry ahead of the highway car in the rightmost lane, which agrees to slow down to create space.
\item \textbf{Negotiation-Highway-Exit}: The exiting car in the leftmost lane requests the adjacent lead vehicle to slow down, allowing for a safe exit maneuver.
\end{enumerate}

Despite these demonstrated successes, several common failure modes emerge:
\begin{enumerate}
    \item \textbf{Stagnation from Model Artifacts}: The distilled model could frequently stall in the \textcolor{myblue}{\texttt{perception-overtake}} scenarios, repeatedly instructing the overtaking car to “hold” while over-assessing the environment (\textcolor{myblue}{\texttt{$*$-safe-distill-failure.mp4}}). We hypothesize this arises from open-loop behavior cloning and reduced exploration during distillation, a limitation that may be mitigated using corrective techniques like DAgger. Notably, this issue is rare in the original \method agent (\textcolor{myblue}{Table~\ref{tab:experiment_cooperative_perception} and \texttt{$*$-safe-debrief-success.mp4}}). 
    Another issue with the distilled model is that it is mechanically memorizing the large model's behavior without thinking, so the smoothness of the trajectories generated by the distilled model is inferior to the large models' (\textcolor{myblue}{\texttt{perception-red-light-safe-distill-success.mp4} and \texttt{perception-red-light-safe-debrief-success.mp4}}).
    
    \item \textbf{Persistent Misinformation in the Communication Network}: For example, in \textcolor{myblue}{\texttt{perception-red-light}} scenarios, once the truck broadcasts an alert about a red-light violator, the alert can persist in agent dialogues even after the violator has cleared the intersection, as the truck lacks a mechanism to retract or update outdated information without a sensor (\textcolor{myblue}{\texttt{perception-red-light-risky-reflection-failure.mp4}}).
    
    \item \textbf{Unpredictable Instruction Following}: Due to inherent randomness in the OpenAI models, agents occasionally exhibit unpredictable behaviors, deviating from expected cooperative actions despite otherwise sound cooperative knowledge.
    
    \item \textbf{Weak Spatial Reasoning}: Constrained by text-based perception rather than richer sensor inputs, the language models display limited spatial reasoning abilities, leading to inefficiencies or failures—such as in perception-highway-exit. We hypothesize that incorporating multi-modal inputs could substantially strengthen spatial reasoning capabilities.

    \item \textbf{Coordination Failures and Misunderstandings}: These failures are more common in the decentralized methods (\textcolor{myblue}{\texttt{Correction+RAG}} and \textcolor{myblue}{\texttt{Reflection}}). For example, agents may collide due to an inability to establish coordinated behaviors based on one-sided knowledge, as seen in cases of indecisive merging (\textcolor{myblue}{\texttt{negotiation-highway-merge-silent-correction-success-indecisive.mp4}}). Similarly, agents sometimes fail to communicate critical information about occluded areas blocked by another car, leading to misunderstandings and failures (\textcolor{myblue}{\texttt{perception-left-turn-reflection-failure-collision.mp4}}).

\end{enumerate}

\section{Additional Results on Safe or Random Configurations}
In the main paper, we report the performance of the policy in risky scenarios; here, we provide results on \textcolor{myblue}{\textbf{safe}} or \textcolor{myblue}{\textbf{random}} (for \texttt{negotiation-highway-merge} and \texttt{negotiation-highway-exit}) scenarios, varying by scenarios. 

Across the safe and random configurations, we observe several key trends. In the \texttt{Cooperative Perception} scenarios, all methods, including baseline zero-shot, achieve perfect or near-perfect success rates with zero collision rates, reflecting the relatively undemanding nature of these safe conditions. Notably, however, the \texttt{Distillation} model shows overly conservative behavior in \texttt{perception-overtake}, achieving only ~14\% success despite the absence of risk, highlighting a tendency to underact in benign settings. In the \texttt{Negotiation} scenarios, the gap between zero-shot and enhanced methods is more pronounced: while zero-shot alone struggles in \texttt{highway-merge} and \texttt{highway-exit} (with success rates around 55–60\%), methods incorporating \texttt{Reflection}, \texttt{Correction+RAG}, or \texttt{Debrief} substantially improve success rates, particularly when paired with communication, reaching up to ~75–93\%. The distilled and centralized memory models consistently deliver the best performance, exceeding 90\% success in these negotiation tasks. Overall, these results indicate that while the safe or random setups may not challenge the perception-based methods, negotiation tasks still benefit significantly from reflective learning, correction mechanisms, and distilled or centralized strategies, underscoring the importance of coordination even in less risky environments.

\begin{table*}[h!]
    \centering
    \caption{\small {{\texttt{Cooperative Perception}}} Scenarios. \texttt{mean $\pm$ std} over 3 trials, each using 30 evaluation episodes.}
    \resizebox{\textwidth}{!}{
    \begin{tabular}{lcc|cc|cc|cc}
    \toprule
    \multicolumn{3}{c|}{\diagbox{Method}{Scenario}} &
    \multicolumn{2}{c|}{Overtake (Perception)} &
    \multicolumn{2}{c|}{Red Light} &
    \multicolumn{2}{c}{Left Turn} \\
    \midrule
    Name & LLM & Comm & 
    CR (\%) $\downarrow$ & SR (\%) $\uparrow$ & 
    CR (\%) $\downarrow$ & SR (\%) $\uparrow$ & 
    CR (\%) $\downarrow$ & SR (\%) $\uparrow$\\
    \midrule
    Zero-shot & Yes & No &
        0.0 $\pm$ 0.0 & 100.0 $\pm$ 0.0 &                            
        0.0 $\pm$ 0.0 & 100.0 $\pm$ 0.0 &                            
        0.0 $\pm$ 0.0 & 100.0 $\pm$ 0.0 \\                           
    +Reflection & Yes & No &
        0.0 $\pm$ 0.0 & 91.1 $\pm$ 15.4 &                            
        0.0 $\pm$ 0.0 & 100.0 $\pm$ 0.0 &                            
        0.0 $\pm$ 0.0 & 100.0 $\pm$ 0.0 \\                           
    +Correction+RAG & Yes & No &
        0.0 $\pm$ 0.0 & 91.1 $\pm$ 7.7 &                            
        0.0 $\pm$ 0.0 & 100.0 $\pm$ 0.0 &                            
        0.0 $\pm$ 0.0 & 100.0 $\pm$ 0.0 \\                           
    \midrule
    Zero-shot & Yes & Yes &
        0.0 $\pm$ 0.0 & 100.0 $\pm$ 0.0 &                            
        0.0 $\pm$ 0.0 & 100.0 $\pm$ 0.0 &                            
        0.0 $\pm$ 0.0 & 100.0 $\pm$ 0.0 \\                           
    +Reflection & Yes & Yes &
        0.0 $\pm$ 0.0 & 97.8 $\pm$ 1.9 &                            
        0.0 $\pm$ 0.0 & 100.0 $\pm$ 0.0 &                            
        0.0 $\pm$ 0.0 & 100.0 $\pm$ 0.0 \\                           
    +Correction+RAG & Yes & Yes &
        0.0 $\pm$ 0.0 & 100.0 $\pm$ 0.0 &                            
        0.0 $\pm$ 0.0 & 100.0 $\pm$ 0.0 &                            
        0.0 $\pm$ 0.0 & 100.0 $\pm$ 0.0 \\                           
    +Debrief & Yes & Yes &
        0.0 $\pm$ 0.0 & 98.9 $\pm$ 1.9 &                            
        0.0 $\pm$ 0.0 & 100.0 $\pm$ 0.0 &                            
        0.0 $\pm$ 0.0 & 100.0 $\pm$ 0.0 \\                           
    \midrule
    Distillation & Yes & Yes &
        0.0 $\pm$ 0.0 & 14.4 $\pm$ 2.2 &                            
        0.0 $\pm$ 0.0 & 100.0 $\pm$ 0.0 &                            
        0.0 $\pm$ 0.0 & 100.0 $\pm$ 0.0 \\                           
    Centralized Memory & Yes & Yes &
        0.0 $\pm$ 0.0 & 100.0 $\pm$ 0.0 &                            
        0.0 $\pm$ 0.0 & 100.0 $\pm$ 0.0 &                            
        0.0 $\pm$ 0.0 & 100.0 $\pm$ 0.0 \\                           
    \bottomrule
    \end{tabular}
    }
    \label{tab:experiment_cooperative_perception}
\end{table*} 

\begin{table*}[h]
    \centering
    \caption{\small {{\texttt{Negotiation}}} Scenarios. \texttt{mean $\pm$ std} over 3 trials, each using 30 evaluation episodes.}
    \resizebox{\textwidth}{!}{
    \begin{tabular}{lcc|cc|cc|cc}
    \toprule
    \multicolumn{3}{c|}{\diagbox{Method}{Scenario}} &
    \multicolumn{2}{c|}{Overtake (Negotiation)} &
    \multicolumn{2}{c|}{Highway Merge} &
    \multicolumn{2}{c}{Highway Exit} \\
    \midrule
    Name & LLM & Comm & 
    CR (\%) $\downarrow$ & SR (\%) $\uparrow$ & 
    CR (\%) $\downarrow$ & SR (\%) $\uparrow$ & 
    CR (\%) $\downarrow$ & SR (\%) $\uparrow$\\
    \midrule
    Zero-shot & Yes & No &
        0.0 $\pm$ 0.0 & 100.0 $\pm$ 0.0 &                            
        41.1 $\pm$ 23.2 & 58.9 $\pm$ 23.2 &                            
        43.3 $\pm$ 11.7 & 56.7 $\pm$ 11.7\\                            
    +Reflection & Yes & No &
        0.0 $\pm$ 0.0 & 100.0 $\pm$ 0.0 &                            
        15.0 $\pm$ 23.2 & 85.0 $\pm$ 23.2 &                            
        38.3 $\pm$ 5.0 & 61.7 $\pm$ 5.0\\                            
    +Correction+RAG & Yes & No &
        0.0 $\pm$ 0.0 & 100.0 $\pm$ 0.0 &                            
        8.9 $\pm$ 3.5 & 91.1 $\pm$ 3.5 &                            
        22.8 $\pm$ 19.2 & 59.4 $\pm$ 11.7\\                            
    \midrule
    Zero-shot & Yes & Yes &
        0.0 $\pm$ 0.0 & 97.8 $\pm$ 1.9 &                            
        45.6 $\pm$ 8.4 & 54.4 $\pm$ 8.4 &                            
        38.9 $\pm$ 2.5 & 60.6 $\pm$ 3.5\\                            
    +Reflection & Yes & Yes &
        0.0 $\pm$ 0.0 & 100.0 $\pm$ 0.0 &                            
        28.9 $\pm$ 8.4 & 70.5 $\pm$ 8.4 &                            
        37.2 $\pm$ 17.7 & 61.1 $\pm$ 15.0\\                            
    +Correction+RAG & Yes & Yes &
        0.0 $\pm$ 0.0 & 98.9 $\pm$ 1.0 &                            
        22.8 $\pm$ 14.9 & 77.2 $\pm$ 14.9 &                            
        35.0 $\pm$ 15.9 & 60.0 $\pm$ 10.4\\                            
    +Debrief & Yes & Yes &
        0.0 $\pm$ 0.0 & 100.0 $\pm$ 0.0 &                            
        24.4 $\pm$ 21.7 & 75.5 $\pm$ 21.7 &                            
        25.0 $\pm$ 11.7 & 75.0 $\pm$ 11.7\\                            
    \midrule
    Distillation & Yes & Yes &
        0.0 $\pm$ 0.0 & 100.0 $\pm$ 0.0 &                            
        6.7 $\pm$ 1.9 & 93.3 $\pm$ 1.9 &                            
        22.8 $\pm$ 2.0 & 77.2 $\pm$ 2.0\\                            
    Centralized Memory & Yes & Yes &
        0.0 $\pm$ 0.0 & 100.0 $\pm$ 0.0 &                            
        9.4 $\pm$ 3.4 & 90.6 $\pm$ 3.4 &                            
        25.5 $\pm$ 2.0 & 74.4 $\pm$ 2.0\\                            

    \bottomrule
    \end{tabular}
    }
    \label{tab:experiment_negotiation}
\end{table*} 

\section{Code Structure Reading Guide}
Our code is available at \url{https://anonymous.4open.science/r/talking-vehicles}

The main implementation is structured as follows:


\dirtree{%
 .1 project/.
 .2 run.sh \textcolor{blue}{\# main entry point}.
 .2 utils/ \textcolor{blue}{\# partial observability utils}.
 .3 partial\_observable\_captioner.py.
 .3 rollout.py.
 .3 shapely\_geometry.py.
 .3 ....
 .2 eval/.
 .3 eval.py.
 .3 ....
 .2 train/.
 .3 train\_centralized\_llm.py.
 .3 train\_decentralized\_llm.py.
 .3 train\_distillation.py.
 .3 ....
 .2 agent/ \textcolor{blue}{\# interface between simulator and policy}.
 .3 agent\_wrapper.py.
 .3 atomic\_command\_follower.py.
 .3 comm\_agent.py.
 .3 comm\_only\_agent.py.
 .3 control\_only\_agent.py.
 .3 local\_planner.py.
 .3 human\_agent.py.
 .3 ....
 .2 policies/ \textcolor{blue}{\# policy and policy learning}.
 .3 llm\_policy.py.
 .3 comm\_gpt\_policy.py.
 .3 dilu\_policy.py.
 .3 ....
 .2 envs/ \textcolor{blue}{\# defines scenario dynamics and agents}.
 .3 multiagent\_env.py.
 .3 scenarios/.
 .4 multiagent\_scenario.py.
 .4 multiagent\_scenarios/.
 .5 highway\_merge/.
 .6 highway\_merge.py.
 .6 highway\_merge\_\_negotiation\_comm\_risky.yaml.
 .6 highway\_merge\_\_negotiation\_comm\_random.yaml.
 .6 ....
 .5 ....
 .4 ....
 .3 ....
 .2 comm/ \textcolor{blue}{\# agent communication module}.
 .3 channel.py.
 .3 transceiver.py.
 .3 mediator.py.
 .3 ....
 .2 wrappers/.
}

\section{Scenarios Set Up and Agent Definition}
We provide an example yaml configuration of a scenario here to demonstrate how a scenario is defined. For a full list of the scenarios set up, please refer to:
\dirtree{%
 .1 project/.
 .2 envs/.
 .3 multiagent\_env.py.
 .3 scenarios/.
 .4 multiagent\_scenario.py.
 .4 multiagent\_scenarios/.
 .5 highway\_merge/.
 .6 highway\_merge.py \textcolor{blue}{\# defines scenario dynamics}.
 .6 highway\_merge\_\_negotiation\_comm\_risky.yaml \textcolor{blue}{\# defines agent type, agent task and scenario parameters}..
 .6 highway\_merge\_\_negotiation\_comm\_random.yaml.
 .6 ....
}

\begin{minted}[label={highway\_merge\_\_negotiation\_comm\_random}]{yaml}
scenarioname: 'HighwayMerge'
map: 'Town06'
accident_prone: False
trigger_point: {x: 196.4, y: 76.4, z: 0.0}
route_var_name: 'highway_merge'
weather:
  sun_altitude_angle: 70
  cloudiness: 5
spectator_transform: {x: 220.0, y: 52.0, z: 50.0, yaw: 0.0}
scenario_knowledge: {
  road_description: {
    25: 'the highway approaching the merge junction',
    29: 'the highway after merge',
    45: 'merging on-ramp',
    741: 'mergeing on-ramp at the highway merge junction',
    16: 'the highway before merge point',
    28: 'the highway after merge',
    735: 'the highway near the highway merge junction',
  },
  junction_description: {
    728: 'a highway merge junction',
  },
  road_geometry: 'highway merge',
  equivalent_roads: {
    'highway': [735, 25, 28, 16],
    'merge ramp': [45, 741],
  },
  available_actions: ['go', 'speed up', 'slow down', 'stop']
}
vehicles:
  - name: 'car1'
    model: 'vehicle.lincoln.mkz_2020'
    scenario: 'highway_merge'
    rolename: 'hero'
    is_focal: True
    agent_type: 'comm_agent'
    sensors: {}
    start: {x: 196.4, y: 76.4, z: 0, roll: 0.0, pitch: 0.0, yaw: -30.0}
    target: {x: 250.0, y: 52.2, z: 0, roll: 0.0, pitch: 0.0, yaw: 0}
    route: [
            {x: 196.4, y: 76.4, z: 0, roll: 0.0, pitch: 0.0, yaw: -30.0},
            {x: 236.9, y: 55.2, z: 0, roll: 0.0, pitch: 0.0, yaw: 0},
            {x: 240.0, y: 52.2, z: 0, roll: 0.0, pitch: 0.0, yaw: 0},
            {x: 250.0, y: 52.2, z: 0, roll: 0.0, pitch: 0.0, yaw: 0}
           ]
    task: 'merge onto the highway (on your left) and you are in a hurry. '
    scenario_knowledge: ${scenario_knowledge}
  - name: 'car2'
    model: 'vehicle.audi.tt'
    scenario: 'highway_merge'
    rolename: 'hero'
    is_focal: True
    agent_type: 'comm_agent'
    sensors: {}
    start: {x: 190.3, y: 52.2, z: 0, roll: 0.0, pitch: 0.0, yaw: 0.0}
    target: {x: 250.0, y: 52.2, z: 0, roll: 0.0, pitch: 0.0, yaw: 0.0}
    route: [
            {x: 190.3, y: 52.2, z: 0, roll: 0.0, pitch: 0.0, yaw: 0.0},
            {x: 250.0, y: 52.2, z: 0, roll: 0.0, pitch: 0.0, yaw: 0.0}
           ]
    task: 'keep on the original highway lane and go forward and you are in a hurry. '
    scenario_knowledge: ${scenario_knowledge}
other_actors:
  - name: 'car3'
    model: 'vehicle.audi.tt'
    scenario: 'highway_merge'
    rolename: 'scenario'
    is_focal: False
    agent_type: 'behavior_agent'
    start: {x: 184.3, y: 52.2, z: 0, roll: 0.0, pitch: 0.0, yaw: 0.0}
    target: {x: 250.0, y: 52.2, z: 0, roll: 0.0, pitch: 0.0, yaw: 0.0}
\end{minted}